\documentclass[]{ummreport}

\usepackage{amsmath,amsfonts,bm}

\def\eqref#1{equation~\ref{#1}}

\def\1{\bm{1}}

\DeclareMathAlphabet{\mathsfit}{\encodingdefault}{\sfdefault}{m}{sl}
\SetMathAlphabet{\mathsfit}{bold}{\encodingdefault}{\sfdefault}{bx}{n}

\usepackage{url}
\usepackage{capt-of}
\usepackage{amsmath,amssymb}
\usepackage{colortbl}
\usepackage{adjustbox}
\usepackage{wrapfig}
\usepackage{marvosym}
\makeatletter
\def\ttl@gobblecontents#1#2#3#4{\ignorespaces}
\makeatother
\usepackage{titletoc}
\titlecontents{lsection}[1.8em]{\addvspace{3pt}\bfseries\sffamily}{\contentslabel{1.8em}}{\hspace*{-1.8em}}{\titlerule*[0.6pc]{.}\contentspage}
\titlecontents{lsubsection}[4.2em]{}{\contentslabel{2.4em}}{\hspace*{-2.4em}}{\titlerule*[0.6pc]{.}\contentspage}
\newcommand{\method}{\mbox{UMM-Reflection}}

\definecolor{bakerlrow}{HTML}{EAF4F0}

\definecolor{ziqigreen}{RGB}{0,100,0}
\newcommand{\ziqi}[1]{}

\newcommand{\GenEvalGainSFT}{12.05}
\newcommand{\WiseGainSFT}{10.97}
\newcommand{\OneIGGainSFT}{3.48}
\newcommand{\CompBenchGainSFT}{4.63}

\newcommand{\OurRepairRate}{64.94}
\newcommand{\SFTRepairRate}{20.59}
\newcommand{\PositionGainSFT}{42.00}
\newcommand{\CountingGainSFT}{10.00}
\newcommand{\BindingGainSFT}{14.00}

\newcommand{\dlt}[1]{\,{\scriptsize\textcolor{gray}{(#1)}}}

\newcommand{\InsertPageThreeHighLevel}{%
  \begin{figure}[!t]
    \centering
    \includegraphics[width=\textwidth]{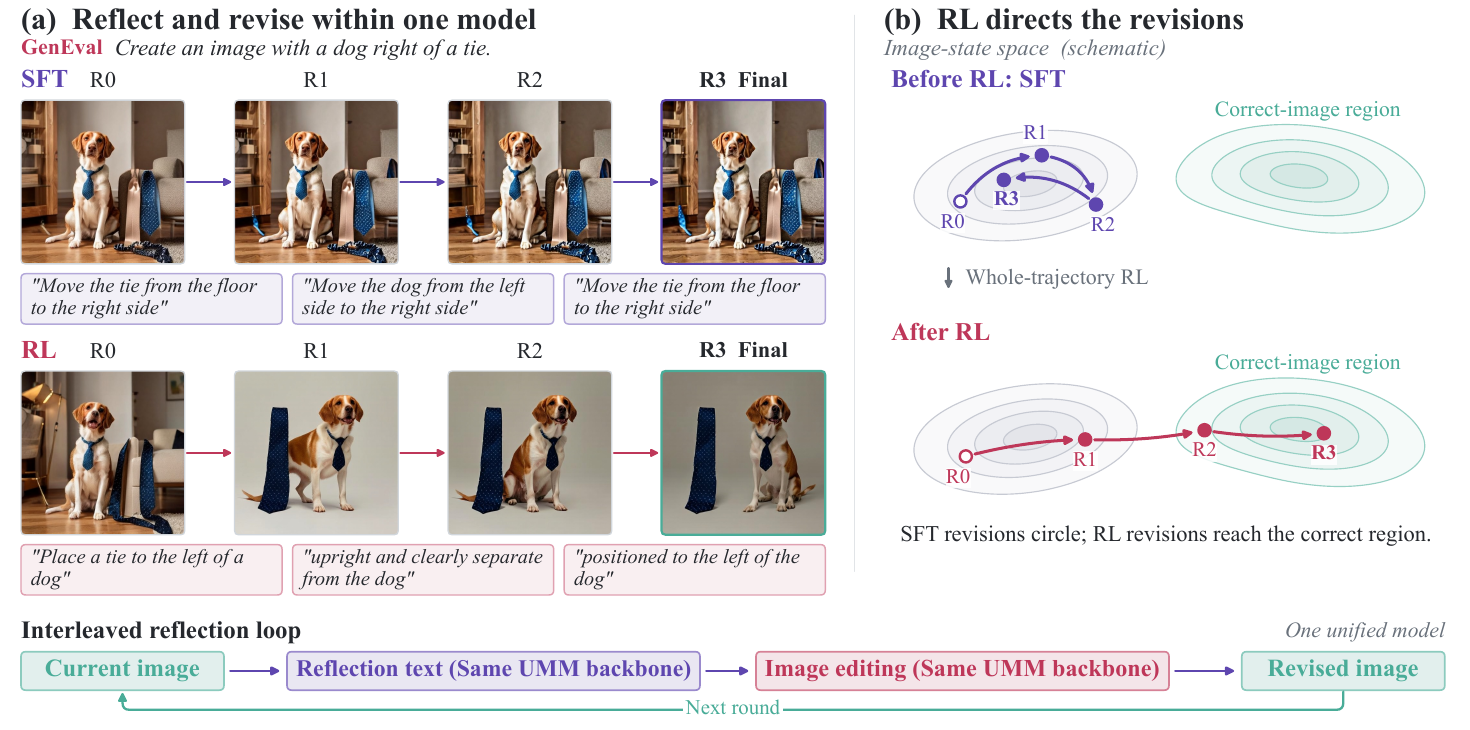}
    \caption{\textbf{Native reflection before and after RL.}
    Left: SFT and RL on the same prompt and seed.
    SFT already produces meaningful revisions (16 SFT rollouts contain a
    correct repair for 78\% of failing training images;
    Appendix~\ref{app:passk}), but one trajectory often circles, as here
    (tie, dog, tie). RL concentrates the policy on revisions that reach the
    correct region (right; measured in Figure~\ref{fig:distribution}).}
    \label{fig:highlevel}
  \end{figure}}

\title{Learning Native Reflection in Unified Models with Interleaved Reinforcement Learning}

\author[1,*,\dagger]{Yijia~Fan}
\author[1,*]{Ziqi~Huang}
\author[1]{Zhongang~Cai}
\author[2]{Yan~Li}
\author[2]{Zimo~Wen}
\author[3]{Wanqi~Yin}
\author[1]{Haiwen~Diao}
\author[1,\ddagger]{Ziwei~Liu}

\affiliation[1]{Nanyang Technological University}
\affiliation[2]{Shanghai Jiao Tong University}
\affiliation[3]{The University of Tokyo}
\contribution[*]{Equal contribution}
\contribution[\dagger]{Work done during an internship at NTU}
\contribution[\ddagger]{Corresponding author}

\abstract{
Unified multimodal models can both look at and render images, so in
principle they can repair their own generations: diagnose what an image gets
wrong, revise it, observe the result, and diagnose again. Whether a revision
helps is known only after it is rendered, so the reflection text and the
image generation must be learned jointly, over the whole loop. Supervised
fine-tuning (SFT) on reflection trajectories gives a cold start but does not
find the high-success repair paths, and naive RL that optimizes
only the renderer or only one head leaves most of the gain untapped. We
introduce \textbf{\method{}}, which applies reinforcement learning (RL) to
complete reflection trajectories inside one unified model: sibling
trajectories share one initial image, so the group-relative advantage
compares reflection strategies, and one trajectory-level advantage updates
both the reflection tokens and the flow-based revisions, avoiding the
combinatorial blow-up of per-round credit assignment. Unlike single-round
editing or pipelines with an external critic, credit flows across rounds and
to both roles of the same model, and no verifier is needed at inference. On
BAGEL, \method{} improves GenEval by \GenEvalGainSFT{} points over SFT, and
the gains transfer to WISE (+\WiseGainSFT{}), OneIG-Bench
(+\OneIGGainSFT{}), and T2I-CompBench++ (+\CompBenchGainSFT{}), none of
which is used in training.
}

\metadata[
\raisebox{-0.20em}{\includegraphics[width=0.025\linewidth]{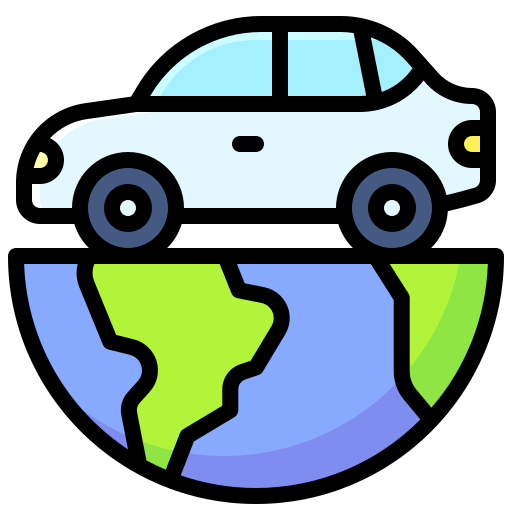}}~~Project Page]
{\href{https://waltstephen.github.io/UMM-Reflection/}
{\texttt{https://waltstephen.github.io/UMM-Reflection}}\\[-2.0ex]}

\metadata[
\raisebox{-0.18em}{\includegraphics[width=0.025\linewidth]{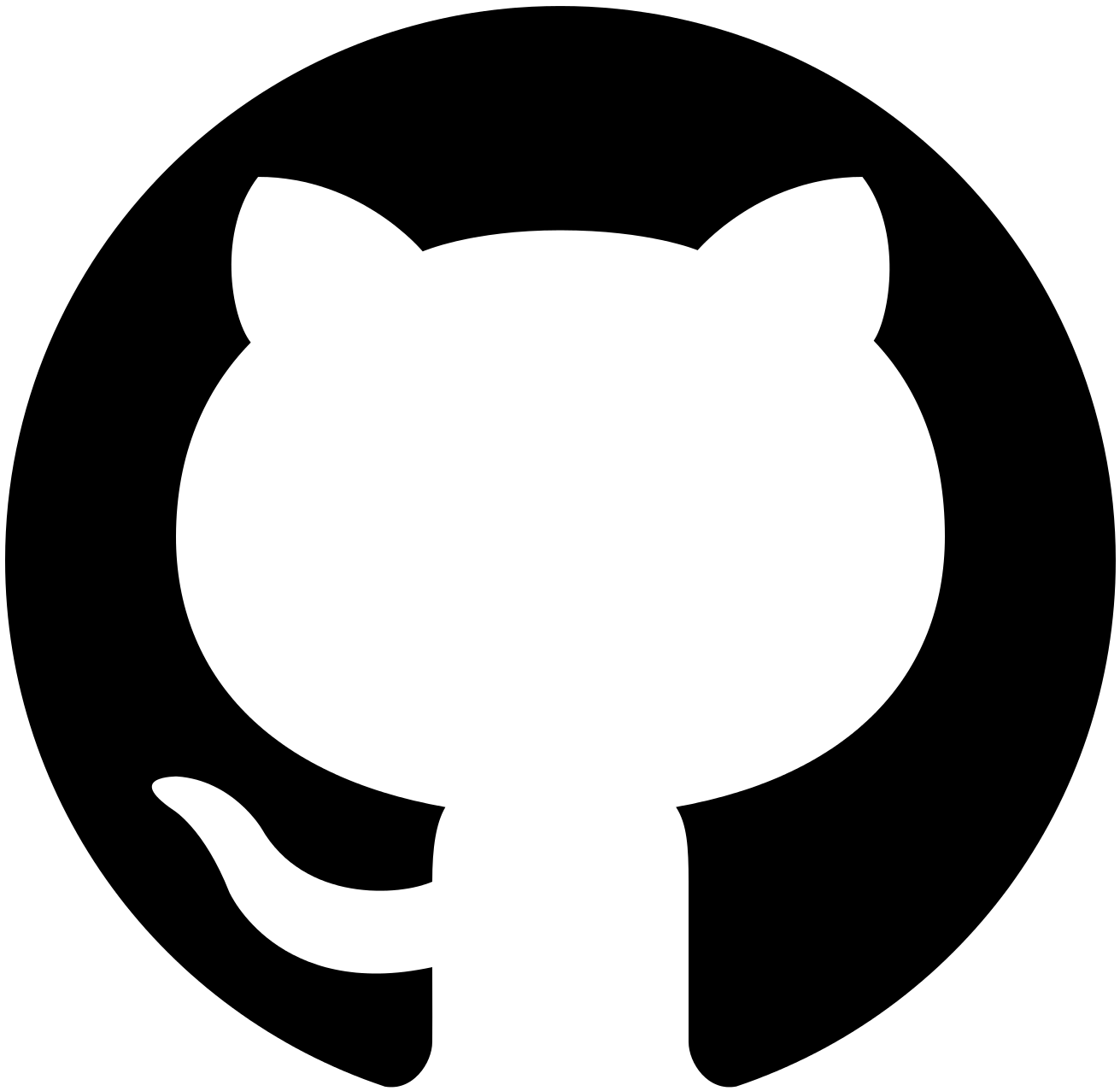}}~~GitHub Repo]
{\href{https://github.com/waltstephen/UMM-Reflection}
{\texttt{https://github.com/waltstephen/UMM-Reflection}}\\[-2.0ex]}

\metadata[
\raisebox{-0.20em}{\includegraphics[width=0.026\linewidth]{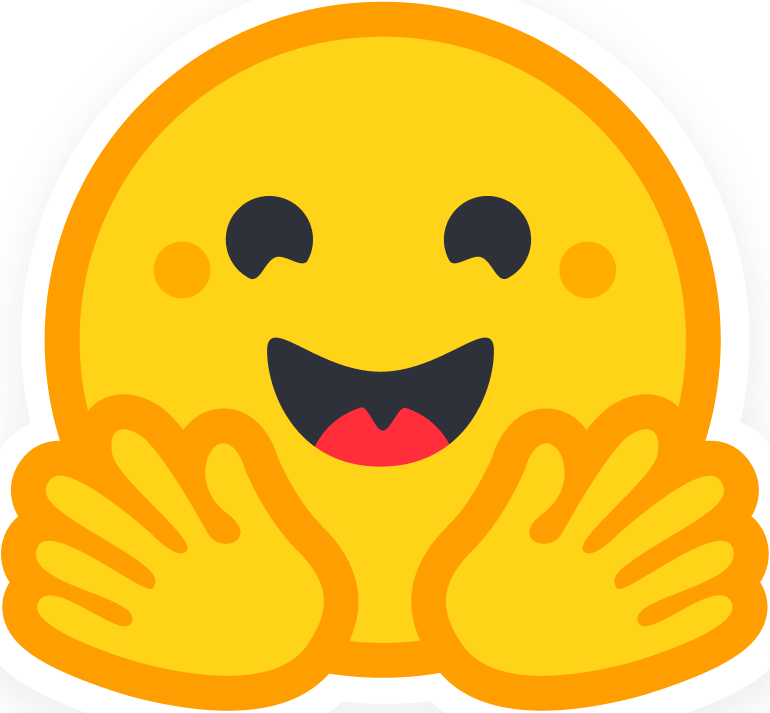}}~~HuggingFace Models \& Data]
{\href{https://huggingface.co/collections/YijiaFan/umm-reflection-6ab95afe909092518d70a158}
{\texttt{https://huggingface.co/collections/YijiaFan/umm-reflection}}\\[-2.0ex]}

\metadata[
\raisebox{-0.16em}{\includegraphics[width=0.026\linewidth]{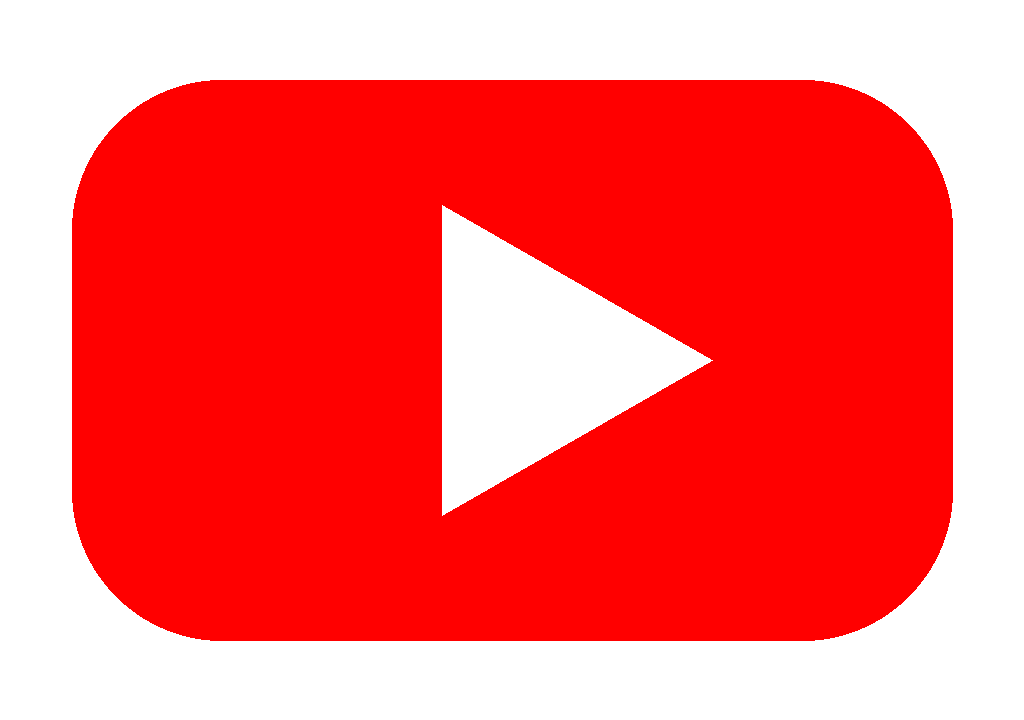}}~~Video Demo]
{\href{https://www.youtube.com/watch?v=YRfpcs4pm-s}
{\texttt{https://www.youtube.com/watch?v=YRfpcs4pm-s}}\\[-2.0ex]}

\metadata[
\raisebox{-0.05em}{\Letter}~~Correspondence]
{Ziwei Liu,~\href{mailto:ziwei.liu@ntu.edu.sg}{\texttt{ziwei.liu@ntu.edu.sg}}}

\begin{document}
\raggedbottom
\maketitle

\section{Introduction}
\label{sec:intro}


As instructions grow compositional, a single render often carries flaws
\citep{geneval2023,compbench2025} that a one-shot text-to-image model
cannot notice, let alone fix.
Unified multimodal models place visual understanding and image generation
in one network \citep{showo2024,emu32024,janus2024,bagel2025}: the model
that renders an image can also look at it.
This enables the loop of inspection, diagnosis, and revision that language
models use for self-correction, beyond step-by-step reasoning
\citep{cot2022,deepseekr12025,selfrefine2023,score2025}: in principle, a
unified model can find the flaws in its own image and render the fix;
\method{} trains it to do so.

\InsertPageThreeHighLevel


Work on unified models approaches this loop from two sides.
One line applies reinforcement learning, but to a single render:
T2I-R1 and ReasonGen-R1 apply GRPO to a textual plan and the single image
rendered from it \citep{t2ir12025,reasongenr12025}, and UniRL turns the model's
understanding of its finished image into a reward for generation
\citep{unirl2025}. The model never revises what it rendered.
The other line lets the model inspect an intermediate image and continue,
learned by supervised imitation of multi-round reasoning-and-editing
trajectories \citep{unit2026,thinkmorph2025}.
Imitation gives these models a cold start, but it does not ensure
that a reflection leads to an effective correction, a gap we quantify in
Section~\ref{sec:study}.
What is missing is reinforcement learning over a unified model's own
multi-round reflection: a signal that credits each reflection for the
visual improvement it produces, rather than for matching a demonstration.
Attaching RL naively does not supply it: optimizing only the renderer, or
only one head of the loop, leaves most of the gain untapped
(Section~\ref{sec:ablation}).


We call this \emph{native reflection}: multi-round inspect--diagnose--revise
behavior carried out by the unified model itself.
Native reflection is what makes the loop trainable as a whole.
The reflection and the revision it triggers come from the same parameters:
the text head writes the diagnosis, and the flow head renders the next
image conditioned on it.
One outcome reward can therefore update both heads along the same
trajectory: a shared outcome-driven advantage reaches every textual and
visual action of the trajectory, and the renderer is trained on the
instructions it actually receives.
This is also what separates the problem from single-round editing and from
pipelines with an external critic: the value of a reflection is known only
after the image it triggers is rendered, and often only after further
rounds, so credit must flow across the whole trajectory and to both the
diagnosis and the rendering. What remains is behavioral: the model must
learn to turn a diagnosis of its own image into a generation action that
fixes it.


We learn native reflection with \method{}.
SFT first teaches the interleaved protocol and meaningful revisions: in
each round the model
examines its current image, writes a reflection, and either generates a
revised image or stops.
Whole-trajectory RL then optimizes complete reflection sequences with two
design choices. All $K$ sibling trajectories start from one shared initial
image, so the group-relative advantage \citep{deepseekmath2024} compares
reflection strategies rather than lucky first draws. One outcome-driven
advantage per trajectory then updates both the reflection tokens and the
flow transitions, so the model learns \emph{which reflections lead to
better images} without the $K^N$ rollouts that per-round credit would
require or a learned value model.
The training-time verifier is never consulted at inference.
Figure~\ref{fig:highlevel} contrasts SFT and RL revisions of the same
image.


The central finding separates \emph{producing useful revisions} from
\emph{reliably choosing them}.
SFT learns more than the format (95\% of its trajectories follow the
protocol; Table~\ref{tab:stages}): its rollouts already contain correct
repairs (Figure~\ref{fig:highlevel}). Yet a single SFT trajectory repairs
only \SFTRepairRate\% of initially incorrect images; after RL, the
conditional repair rate rises to \OurRepairRate\% (Table~\ref{tab:ablation}).
This gap translates to substantial accuracy gains on GenEval
\citep{geneval2023} (+\GenEvalGainSFT{} over SFT) and WISE
\citep{wise2025} (+\WiseGainSFT{}).
Base, SFT, and RL start from nearly identical single-round accuracy
(70--73): the first image receives no RL loss, and the gains come from the
reflection rounds.
Updating the generator alone with the same number of RL updates (direct T2I-RL) lifts
single-shot GenEval from 71 to 76 but does not transfer (WISE 54 versus 55
for Base), whereas \method{} reaches 84 and 74.
Gains also transfer to OneIG-Bench \citep{oneig2025} and T2I-CompBench++
\citep{compbench2025}, neither seen in training.
Representation analysis (Section~\ref{sec:study}) shows that RL leaves the
model's perception and internal correctness readout nearly unchanged; among
the revisions SFT already produces, it selects those that move a failing
image into the region this readout marks as correct
(Figure~\ref{fig:distribution}), finding better repair paths rather than
creating a new capability.


We summarize our contributions as follows:
\begin{itemize}
\item We enable reinforcement learning for multi-round reflection in
      unified models: one whole-trajectory advantage, computed over siblings
      that share one initial image, jointly optimizes the textual
      reflections and the flow-based revisions of the same model without
      per-round branching.
\item We show that imitation already teaches meaningful revisions but
      applies them unreliably, and that RL makes them reliable by selecting
      repair paths the backbone already has.
\item \method{} improves over reflection SFT on four benchmarks while
      training on one; ablations show that neither direct RL on the
      generator nor Best-of-4 selection matches it.
\end{itemize}

\section{Related Work}
\label{sec:related}


\paragraph{Self-correction loops, in text and in pixels.}
Self-Refine and Reflexion let a language model critique and revise its own
draft by prompting alone \citep{selfrefine2023,reflexion2023}, yet without
external feedback such intrinsic self-correction can lower accuracy
\citep{cannotselfcorrect2023}. SCoRe traces this to offline correction traces
and shows that online multi-turn RL on the model's own attempts is needed
\citep{score2025,rise2024}. Image generation has adopted the loop
but not this lesson: Idea2Img, iterative refinement, ReflectionFlow, SLD, and
GenArtist pair an external critic with a separate
renderer \citep{idea2img2023,iterativerefine2026,reflectionflow2025,sld2024,genartist2024}.
The critic sees only pixels, neither model is optimized against the other,
and the critic stays online at inference. We train both roles as one policy
under one outcome reward and drop the verifier at inference.


\paragraph{Reinforcement learning for visual generators.}
DDPO and DPOK optimize the denoising chain with policy gradients
\citep{ddpo2023,dpok2023}, ImageReward and Diffusion-DPO learn from human
preferences \citep{imagereward2023,diffusiondpo2023}, and Flow-GRPO and
DanceGRPO bring group-relative optimization \citep{deepseekmath2024} to
flow-matching generators \citep{flowgrpo2025,dancegrpo2025}. In each, the
policy is a single prompt-to-image pass that never observes its own render.
We place Flow-GRPO's rendering transitions inside a multi-round trajectory
whose single advantage credits both the reflection tokens and the renders
they trigger.


\paragraph{Reasoning and reflection in unified generators.}
Unified models share one network for understanding and generation via
discrete tokens \citep{chameleon2024,emu32024,showo2024}, decoupled
visual encoders \citep{janus2024,januspro2025}, text with a diffusion or flow
decoder \citep{transfusion2024,bagel2025}, or bridging queries and tokens
\citep{metaqueries2025,blip3o2025,bividgen2026}, and post-training further
couples their understanding with generation \citep{reca2026,alphagrpo2026}. One line improves a
single render: T2I-R1 and ReasonGen-R1 apply GRPO to a textual plan and its
image \citep{t2ir12025,reasongenr12025}, UniRL rewards generation with the
model's own answers about its finished image \citep{unirl2025}, and PARM and GoT
verify or structure the generation process \citep{parm2025,got2025}; none revises the image. A second line
(Thinking with Generated Images, MINT, Uni-CoT, IRG, ThinkMorph, UniT)
inspects an intermediate image and continues
\citep{thinkinggenimages2025,mint2025,unicot2025,irg2025,thinkmorph2025,unit2026},
but is trained mainly by imitating synthesized trajectories, which, as SCoRe
predicts and Section~\ref{sec:study} measures, gives a cold start
without the high-success repair paths. Multi-round regeneration has also been
trained with RL, but as separate rounds that do not inherit earlier context
\citep{r32026}. \method{} applies outcome-driven RL
to complete inspect-and-revise trajectories in one unified policy.

\section{Methodology}
\label{sec:method}

We formulate native reflection as a policy that repeatedly inspects,
diagnoses, and revises its own image within a single unified model.
This section describes the reflection protocol (\S\ref{sec:protocol}),
the trajectory data used to initialize it (\S\ref{sec:data}), the
supervised cold-start (\S\ref{sec:sft}), and the whole-trajectory RL
stage that turns this cold start into effective repair
(\S\ref{sec:rl}).

\begin{figure}[t]
\centering
\includegraphics[width=\linewidth]{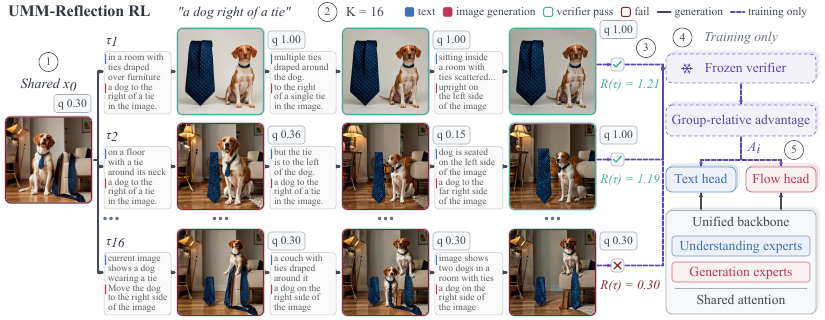}
\caption{\textbf{\method{} RL.} (1) $K{=}16$ rollouts share one detached
initial image $x_0$. (2) Each interleaves the model's own reflection
(verbatim) with its renders for up to three rounds. (3) A frozen verifier
scores every image ($q$) and trajectory ($R(\tau)$). (4) Group normalization
gives one advantage $A_i$ per trajectory, (5) which updates both the text
and flow heads. Green/red frames: verifier pass/fail; dashed: training only.}
\label{fig:pipeline}
\end{figure}

\subsection{Reflection protocol}
\label{sec:protocol}

For a request $c$, the model first produces an image $x_0$.
At each round~$t$, the model observes the request, the text--image
history, and the current image $x_t$, and emits a structured reflection:
\[
u_t \;\sim\; \pi_{\theta}^{\text{text}}
  (\,\cdot \mid c,\, x_{\leq t},\, u_{<t}),
\qquad
a_t \in \{\textsc{edit},\;\textsc{done}\}.
\]
An \textsc{edit} action carries a natural-language correction $e_t$;
the same model then renders
$x_{t+1}\sim\pi_{\theta}^{\text{flow}}(\cdot\mid c, x_t, e_t)$.
A \textsc{done} action returns the current image.
The loop runs for at most three repair rounds in the primary experiments.

Each reflection is a tagged response whose main fields are
\texttt{[THINKING]}, \texttt{[ACTION]}, and \texttt{[EDIT]}
(full format in Appendix~\ref{app:sftdetails}); verifier scores never
enter the policy observation.
Figure~\ref{fig:pipeline} shows the trajectory structure.

\subsection{Trajectory data construction}
\label{sec:data}

Learning the protocol requires multi-round inspect--diagnose--revise
trajectories, which the target model cannot yet produce. We generate them
with external models and distill them into the unified model via SFT.
GPT-5.5 acts as the critic: it turns each request into verifiable
constraints, inspects each image, writes a structured reflection, and
issues one atomic edit instruction or a \textsc{done} verdict. Qwen-Image
renders the initial image and Qwen-Image-Edit executes each edit; BAGEL
takes no part, so its own failure modes are not distilled back into the
supervision.
The $29{,}529$ accepted trajectories are of three types:
\emph{one-shot} ($9{,}000$), where the initial image already satisfies
all constraints; \emph{natural-repair} ($8{,}645$), where a genuinely
failing initial image is fixed in one or two rounds without injected
corruption; and \emph{planned-progression} ($11{,}884$), where a complex
request is fulfilled over two or three ordered milestones. Prompts come
from Puffin-4M, Poster100K, OmniEdit, AnyEdit, and GEdit-Bench, with no
overlap with any evaluation benchmark.

\subsection{Supervised initialization}
\label{sec:sft}

SFT teaches the interleaved protocol on the trajectories from
\S\ref{sec:data} with an autoregressive loss on reflection text and a
flow-matching loss on each edited image,
$\mathcal{L}_{\text{SFT}} = \mathcal{L}_{\text{AR}} + \lambda_{\text{img}}\,\mathcal{L}_{\text{FM}}$,
starting from the base BAGEL checkpoint for one epoch
(details in Appendix~\ref{app:sftdetails}).
This stage supplies a consistent interface between diagnosis and
corrective generation; the subsequent RL stage starts from this
SFT checkpoint.

\subsection{Whole-trajectory reinforcement learning}
\label{sec:rl}

SFT teaches the model to produce well-formed reflections, but
well-formed text does not guarantee effective repair
(\S\ref{sec:study}).
We now describe the RL stage that optimizes for visual outcomes.

\paragraph{Shared-root sampling.}
For each request, we sample one initial image and detach it from the
computation graph.
RL optimizes only the reflection-and-editing rounds that follow; the
initial text-to-image generation receives no policy-gradient
signal.\footnote{BAGEL does not natively support interleaved
text--image generation in a single forward pass. We implement the
multi-round loop through an external controller that feeds each
round's reflection and image back into the model as a new turn,
enforcing the interleaved protocol described in \S\ref{sec:protocol}.}
From those identical root pixels, we sample $K\!=\!16$ complete
reflection trajectories $\{\tau_i\}_{i=1}^{K}$, each running until
its own \textsc{done} action or the repair cap.
We do not prune siblings, retain only the best intermediate image,
or use best-of-$K$ selection at deployment.

\paragraph{Reward.}
A frozen verifier assigns a graded alignment score
$q_t\in[0,1]$ to each image, aggregated from its own detector outputs
under the official thresholds so that a partial repair yields a nonzero
change (Appendix~\ref{sec:graded}).
Let $\Delta_t = q_{t+1} - q_t$ and $[v]_+ = \max(v,0)$.
The trajectory reward is
\begin{align}
R(\tau)
  &= q_T
   + \alpha \textstyle\sum_{t}[\Delta_t]_+
   + \beta\, S_{\text{multi}}(\tau)
   - \lambda \textstyle\sum_{t}[-\Delta_t]_+
   - p\,\mathbf{1}[\text{premature } \textsc{done}],
   \label{eq:reward} \\[4pt]
S_{\text{multi}}(\tau)
  &= \textstyle\sum_{t}[\Delta_t]_+
   - \max\!\bigl(\{[\Delta_t]_+\}_{t}\cup\{0\}\bigr).
   \label{eq:multi}
\end{align}
The terminal score $q_T$ rewards the final image quality.
The progress terms $[\Delta_t]_+$ reward each round that improves the
image; the damage penalty $\lambda\sum[-\Delta_t]_+$ discourages
regressions.
$S_{\text{multi}}$ adds credit when improvement comes from more than
one edit rather than a single lucky fix; it deliberately favors
trajectories that keep improving across rounds, since multi-round
correction is the behavior we aim to train.
Premature \textsc{done} means stopping when the verifier does not
accept the current image.
We use $\alpha\!=\!\beta\!=\!0.3$ and $\lambda\!=\!p\!=\!0.5$.

\paragraph{Group-relative advantage.}
Within each shared-root group, advantages are
\[
A_i = \operatorname{clip}\!\left(
  \frac{R(\tau_i) - \mu_R}
       {\max(\sigma_R,\, 0.1)},\;
  {-1},\; 1\right).
\]
We assign one advantage per trajectory rather than per round. A
group-relative estimate for each round would require sibling groups at
every round: branching $K$ ways at each of $N$ rounds needs $K^N$ rollouts
per root ($16^3=4{,}096$ for three rounds), which is impractical for an
image-generating policy. A per-round critic, as in PPO, would instead
require training a value model over multi-round image--text states, with
the data scale that entails. The trajectory-level advantage keeps the
$K$-sample cost of GRPO while the per-round progress terms in
$R(\tau)$ still reward each round that improves the image.

\paragraph{Text--flow coordination.}
Let $\mathcal{I}_i^{c}$ denote the policy-active positions for
channel $c \in \{\text{text},\,\text{flow}\}$, and $\rho_{ij}^{c}$
the corresponding likelihood ratio.
The clipped surrogate for each channel is
\[
\mathcal{J}_c =
\mathbb{E}_{i,\,j \in \mathcal{I}_i^{c}}
\min\!\bigl(\rho_{ij}^{c} A_i,\;
\operatorname{clip}(\rho_{ij}^{c},\,
  1{-}\epsilon_c,\,1{+}\epsilon_c)\,A_i\bigr)
- \eta_c \mathcal{K}_c.
\]
The key design choice is that both channels share the \emph{same}
trajectory-level $A_i$: the text policy and the flow renderer are
not normalized separately.
A reflection that leads to a better image raises the advantage for
both the diagnostic tokens and the rendering transitions that followed,
so the model learns which reflections lead to which visual outcomes.
$\mathcal{K}_c$ is a channel-specific KL penalty against the frozen SFT
reference.
Flow transitions use the Flow-GRPO SDE sampler \citep{flowgrpo2025}; per
active repair we train on two contiguous stochastic transitions. For text, credited positions
are sampled policy tokens excluding prompt and formatting.

The verifier and the reference policy are used only during training; at
inference only the unified model runs.

\section{Experiments}
\label{sec:experiments}

\subsection{Setup}
\label{sec:setup}

\paragraph{Training and evaluation.}
We build on BAGEL \citep{bagel2025}, the most widely used open unified
model that both understands and generates images in one network, with
understanding and generation experts that share attention; this lets one
trajectory-level advantage update the reflection text and the renderer of
the same model. RL starts from the reflection-SFT checkpoint (\S\ref{sec:sft}) and samples
from a $3{,}000$-prompt pool over six GenEval families, with two roots and
$K\!=\!16$ siblings per update and 20 training denoising steps; unless
otherwise specified, RL runs for $1{,}000$ updates. We evaluate one image per
prompt with 50 denoising steps at $512^2$ and at most three repairs, using
the same checkpoint on GenEval (all 553 official prompts, unfiltered), WISE (1,000), OneIG-Bench (OneIG; 695 alignment prompts),
and T2I-CompBench++ (2,400); protocol and scorer details are in
Appendix~\ref{app:protocol}.

\paragraph{Baselines.}
The main comparison uses the same BAGEL backbone throughout:
\emph{Base}, unmodified BAGEL, single-pass generation at $512^2$, and
\emph{SFT}, the reflection-supervised parent, producing multi-round
trajectories without RL.
Ablation baselines that remove or replace one ingredient are defined in
Section~\ref{sec:ablation}.

\subsection{Main results}
\label{sec:mainresults}

\begin{table}[t]
\centering
\caption{\textbf{GenEval: all six compositional requirements.}
Native $0$--$1$ scores. Res.: image side (\textit{n/r}: not reported);
$\dagger$: reported in another model's paper. Published settings differ
from our one-image, instruction-voice evaluation; e.g., BAGEL reports 0.82
under its native protocol \citep{bagel2025}, whereas the local rows use our
protocol (Appendix~\ref{app:protocol}). Bold: best in the local
block.}
\label{tab:geneval}
\fontsize{9}{10.5}\selectfont
\setlength{\tabcolsep}{5pt}
\renewcommand{\arraystretch}{1.04}
\begin{adjustbox}{max width=\linewidth}
\begin{tabular}{@{}lrccccccc@{}}
\toprule
Model & Res. & Single & Two & Count & Colors & Position & Binding & Overall \\
\midrule
\multicolumn{9}{l}{\textit{Reported results: native settings and original reporting precision}} \\
PixArt-$\alpha$$^{\dagger}$~\citep{showo2024} & 512 & 0.98 & 0.50 & 0.44 & 0.80 & 0.08 & 0.07 & 0.48 \\
SD2.1~\citep{geneval2023} & 768 & 0.98 & 0.51 & 0.44 & 0.85 & 0.07 & 0.17 & 0.50 \\
DALL$\cdot$E 2$^{\dagger}$~\citep{showo2024} & 1024 & 0.94 & 0.66 & 0.49 & 0.77 & 0.10 & 0.19 & 0.52 \\
Emu3-Gen~\citep{emu32024} & 512 & 0.98 & 0.71 & 0.34 & 0.81 & 0.17 & 0.21 & 0.54 \\
SDXL~\citep{geneval2023} & 1024 & 0.98 & 0.74 & 0.39 & 0.85 & 0.15 & 0.23 & 0.55 \\
DALL$\cdot$E 3$^{\dagger}$~\citep{januspro2025} & 1024 & 0.96 & 0.87 & 0.47 & 0.83 & 0.43 & 0.45 & 0.67 \\
SD3-Medium$^{\dagger}$~\citep{januspro2025} & \textit{n/r} & 0.99 & 0.94 & 0.72 & 0.89 & 0.33 & 0.60 & 0.74 \\
LWM$^{\dagger}$~\citep{showo2024} & \textit{n/r} & 0.93 & 0.41 & 0.46 & 0.79 & 0.09 & 0.15 & 0.47 \\
SEED-X$^{\dagger}$~\citep{showo2024} & \textit{n/r} & 0.97 & 0.58 & 0.26 & 0.80 & 0.19 & 0.14 & 0.49 \\
TokenFlow~\citep{tokenflow2024} & 256 & 0.97 & 0.66 & 0.40 & 0.84 & 0.17 & 0.26 & 0.55 \\
ILLUME~\citep{illume2024} & 512 & 0.99 & 0.86 & 0.45 & 0.71 & 0.39 & 0.28 & 0.61 \\
Janus~\citep{janus2024} & 384 & 0.97 & 0.68 & 0.30 & 0.84 & 0.46 & 0.42 & 0.61 \\
Show-o-512~\citep{showo2024} & 512 & 0.98 & 0.80 & 0.66 & 0.84 & 0.31 & 0.50 & 0.68 \\
Janus-Pro-7B~\citep{januspro2025} & 384 & 0.99 & 0.89 & 0.59 & 0.90 & 0.79 & 0.66 & 0.80 \\
\midrule
\multicolumn{9}{l}{\textit{Local evaluation: same prompts, resolution, and scorer}} \\
BAGEL-Base & 512 & \textbf{1.00} & 0.89 & 0.63 & 0.82 & 0.47 & 0.48 & 0.71 \\
BAGEL-SFT & 512 & \textbf{1.00} & 0.88 & 0.58 & 0.87 & 0.47 & 0.51 & 0.72 \\
\rowcolor{bakerlrow}\method{} & 512 & 0.95 & \textbf{0.96} & \textbf{0.68} & \textbf{0.90} & \textbf{0.89} & \textbf{0.65} & \textbf{0.84} \\
\bottomrule
\end{tabular}

\end{adjustbox}
\end{table}

\begin{table}[t]
\centering
\caption{\textbf{Transfer to benchmarks unseen in RL training.} Native
$0$--$1$ scale; gray values are changes relative to BAGEL-Base.}
\label{tab:main}
\small
\setlength{\tabcolsep}{5pt}
\begin{adjustbox}{max width=\linewidth}
\providecommand{\TabVal}[2]{\phantom{\dlt{#2}}#1\dlt{#2}}
\providecommand{\TabNoD}[1]{\phantom{\dlt{+0.00}}#1\phantom{\dlt{+0.00}}}
\begin{tabular}{@{}lccc@{}}
\toprule
Model & WISE & OneIG & T2I-CompBench++ \\
\midrule
\multicolumn{4}{l}{\textit{Local evaluation: same prompts, resolution, and scorer}} \\
BAGEL-Base & \TabNoD{0.55} & \TabNoD{0.80} & \TabNoD{0.49} \\
BAGEL-SFT & \TabVal{0.63}{+0.08} & \TabVal{0.79}{\textminus0.01} & \TabVal{0.50}{+0.01} \\
\rowcolor{bakerlrow}\method{} & \TabVal{\textbf{0.74}}{+0.19} & \TabVal{\textbf{0.83}}{+0.02} & \TabVal{\textbf{0.55}}{+0.06} \\
\bottomrule
\end{tabular}

\end{adjustbox}
\end{table}

\begin{figure*}[t]
\centering
\includegraphics[width=\linewidth]{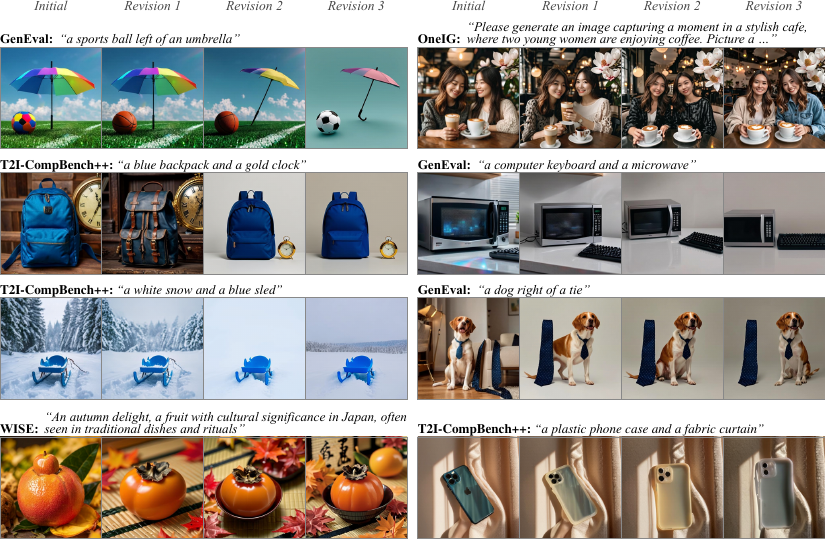}
\caption{\textbf{Case studies across four benchmarks.}
Each row shows the prompt, the initial image, and three reflection-guided
revisions (left to right), with examples from GenEval, WISE, OneIG, and
T2I-CompBench++. The model identifies spatial, color, material, and compositional
errors through its \texttt{[THINKING]} output and issues targeted edits.}
\label{fig:case_study}
\end{figure*}

\paragraph{In-domain results.}
Table~\ref{tab:geneval} places \method{} among reported GenEval results.
Under the same protocol, \method{} reaches 0.84, against 0.71 for
BAGEL-Base and 0.72 for reflection SFT (+12 points over SFT). The gain is
concentrated in the families that require fixing a composition:
\emph{position} rises by +\PositionGainSFT{} points over SFT (0.47 to 0.89),
color \emph{binding} by +\BindingGainSFT{}, and \emph{counting} by
+\CountingGainSFT{}. Paired over prompts, RL beats SFT on 109 prompts and loses on 38
($p<10^{-8}$, McNemar), while on the initial images alone the split is 48--32
($p=0.09$, no significant difference), and the gain is made in the
reflection rounds.

\begin{wrapfigure}[13]{r}{0.40\linewidth}
\vspace{-14pt}
\centering
\includegraphics[width=\linewidth]{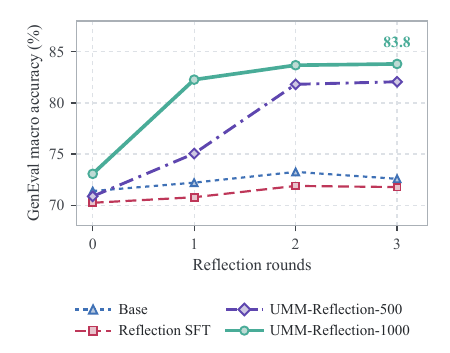}
\vspace{-18pt}
\caption{\textbf{Test-time scaling on GenEval.} Macro accuracy versus
reflection rounds. Other benchmarks: Appendix Figure~\ref{fig:tts4}.}
\label{fig:tts}
\vspace{-12pt}
\end{wrapfigure}
\paragraph{Transfer.}
Table~\ref{tab:main} evaluates the same checkpoint on three benchmarks never
used in RL training. \method{} improves over reflection SFT on all three:
+\WiseGainSFT{} points on WISE, +\CompBenchGainSFT{} on T2I-CompBench++, and
+\OneIGGainSFT{} on OneIG, where SFT alone falls slightly below Base.
Figure~\ref{fig:case_study} shows reflection trajectories on all four
benchmarks. Section~\ref{sec:ablation} isolates the contribution of each ingredient.

\subsection{Multi-round test-time scaling}
\label{sec:tts}

Figure~\ref{fig:tts} shows the per-round GenEval score; the other three
benchmarks follow the same pattern (Appendix Figure~\ref{fig:tts4}).
SFT's three rounds add roughly +2 points on GenEval and flatten after
round~1; the model edits but does not reliably improve.
After RL, round~1 alone adds +9 points, and the model continues to
gain through round~3.
The initial-image accuracy (R0) is comparable across arms (70--73),
confirming that the gap comes from multi-round correction, not a
better first image.

The conditional repair rate rises from \SFTRepairRate\% (SFT) to
\OurRepairRate\% (RL\,1000). Within GenEval families, the largest gains
are on \emph{position} (+34) and \emph{color\_attr} (+17).
Full per-round and per-family statistics are in
Appendix~\ref{app:paired}.

\section{Ablations}
\label{sec:ablation}

\begin{table}[t]
\centering
\caption{\textbf{Ablations on GenEval (553 official prompts).} Six-family macro accuracy
($0$--$100$). Repair and Damage are percentages of initially incorrect
images fixed and initially correct images broken; $\Delta$ is computed
before rounding. All reward variants
share the same SFT parent, prompt pool, and training topology.}
\label{tab:ablation}
\small
\setlength{\tabcolsep}{4pt}
\begin{adjustbox}{max width=\linewidth}
\begin{tabular}{@{}lrrrrrrr@{}}
\toprule
 & R0 & Final & $\Delta$ & Repair & Damage & Edits & Images \\
\midrule
\multicolumn{8}{@{}l}{\textit{Components}} \\
BAGEL-Base & 71 & 71 & 0 & -- & -- & 0 & 1 \\
+ direct RL on the renderer (T2I-RL) & 76 & 76 & 0 & -- & -- & 0 & 1 \\
+ inspect-and-edit loop (Self-Agentic) & 71 & 77 & +5 & 27.6 & 3.6 & 3.00 & 4.0 \\
+ reflection SFT & 70 & 72 & +2 & 20.6 & 6.5 & 1.65 & 2.7 \\
\rowcolor{bakerlrow}+ reflection SFT + trajectory RL (\method{}) & 73 & 84 & +11 & 64.9 & 8.8 & 3.00 & 4.0 \\
direct RL $\to$ reflection SFT $\to$ trajectory RL (500 updates only) & 74 & 81 & +7 & 48.3 & 7.9 & 3.00 & 4.0 \\
\midrule
\multicolumn{8}{@{}l}{\textit{Training length (same run)}} \\
reflection SFT + trajectory RL, 100 updates & 71 & 75 & +3 & 35.6 & 9.5 & 3.00 & 4.0 \\
reflection SFT + trajectory RL, 500 updates & 71 & 82 & +11 & 61.1 & 9.3 & 3.00 & 4.0 \\
\rowcolor{bakerlrow}\method{} (1{,}000 updates) & 73 & 84 & +11 & 64.9 & 8.8 & 3.00 & 4.0 \\
\midrule
\multicolumn{8}{@{}l}{\textit{Head ablation (the other head frozen)}} \\
flow-only RL (frozen text head) & 71 & 73 & +2 & 22.8 & 7.3 & 3.00 & 4.0 \\
text-only RL (frozen flow head) & 71 & 78 & +7 & 49.4 & 10.1 & 3.00 & 4.0 \\
\rowcolor{bakerlrow}joint RL (\method{}) & 73 & 84 & +11 & 64.9 & 8.8 & 3.00 & 4.0 \\
\midrule
\multicolumn{8}{@{}l}{\textit{Same four-image budget}} \\
Best-of-4 T2I-RL, selected by Base UND & 76 & 80 & +4 & -- & -- & 0 & 4 \\
Best-of-4 T2I-RL, selected by \method{} UND & 76 & 80 & +5 & -- & -- & 0 & 4 \\
reflection SFT, forced to edit in all three rounds & 70 & 72 & +2 & 28.8 & 9.7 & 3 & 4 \\
\rowcolor{bakerlrow}\method{} & 73 & 84 & +11 & 64.9 & 8.8 & 3.00 & 4.0 \\
\midrule
\multicolumn{8}{@{}l}{\textit{Reward terms}} \\
\rowcolor{bakerlrow}\method{} & 73 & 84 & +11 & 64.9 & 8.8 & 3.00 & 4.0 \\
without the multi-improvement term ($\beta=0$) & 63 & 79 & +16 & 61.7 & 11.2 & 3.00 & 4.0 \\
\bottomrule
\end{tabular}

\end{adjustbox}
\end{table}

Table~\ref{tab:ablation} isolates each ingredient; the results support
five conclusions.

\paragraph{The gains stack.}
Direct Flow-GRPO on the renderer (T2I-RL, 1{,}000 updates from Base) raises
single-shot accuracy from 71 to 76 but leaves nothing to repair. Forcing the
untuned Base through three inspect-and-edit rounds (Self-Agentic) adds +5
without training. Reflection SFT adds +2; fine-tuning Base on the final images
of the same 29{,}529 trajectories leaves single-pass accuracy on the raw official
prompt at 75, the same as Base, so the SFT images alone do not improve the
generator. Trajectory RL on top of SFT adds +11 and triples the
repair rate from 21\% to 65\%. Placing direct RL before both stages carries
its single-shot advantage through, reaching 81 after 500 updates.

\paragraph{Most of the gain arrives within 500 updates.}
Along the same run, GenEval rises from 72 (SFT) to 75, 79, and 82 after
100, 200, and 500 updates, and reaches 84 at 1{,}000.
The repair rate follows the same path, from 21\% to 61\% at 500 updates
and 65\% at 1{,}000, while damage stays between 8\% and 10\%.
Initial-image accuracy stays at 71--73 throughout, so the gain comes from
the reflection rounds at every checkpoint.

\paragraph{Both heads must be trained.}
Freezing one head while applying the same trajectory RL isolates what each
side contributes. Training only the flow head leaves the model close to SFT
(73, repair rate 22.8\%): a better renderer does not help when the
reflections that drive it do not improve. Training only the text head
recovers most of the gain (78, 49.4\%), so learning what to write is the
larger part. Joint training reaches 84 and 64.9\%, six points above the
best single head: the renderer must also learn to execute the reflections
the text head now writes. This is the joint optimization that a single
unified model makes possible. Holding image, renderer, and noise fixed and
swapping only the instruction confirms that the learned text carries the
repair (Appendix~\ref{app:swap}).

\paragraph{The gain is not best-of-$N$ sampling or extra edits.}
At the same four-image budget, selecting one of four images from the stronger
T2I-RL renderer with a single native understanding call (Best-of-4) reaches 80
with either selector; \method{} reaches 84 (paired McNemar $p\le0.04$).
Forcing SFT to edit in all three rounds leaves it at 72, the same as
unforced SFT.
On the 62 prompts where all four independent Base draws fail, RL reflection
recovers 60\% (SFT: 12\%).

\paragraph{The multi-improvement term matters.}
Removing the multi-improvement term ($\beta\!=\!0$) preserves the
repair rate but degrades the initial image from 73 to 63, also yielding
79.

\section{Study: What Changes Inside the Model}
\label{sec:study}

\subsection{Training dynamics: the interface locks in first, then repair improves}
\label{sec:dynamics}

Appendix Figure~\ref{fig:training} summarizes the 1,000-update RL run.
Protocol compliance converges within the first fifty updates:
invalid trajectories drop below 1\% and stay there.
After that, the reward curve is driven by improving repair quality: 
successful repairs per edit rise steadily from 11\% to 38\%,
while the damage rate on initially correct images falls from 20\% to
10\%. Terminal exactness under the training verifier reaches 81\%.
There is no reward collapse or protocol regression within 1,000
updates.

On the full GenEval test set (553 prompts), SFT's three reflection
rounds add +2 points of macro accuracy; RL raises this to +11,
with the gain concentrated in the repair process rather than the
initial generation (Section~\ref{sec:tts}).

\subsection{What RL changes in the model}
\label{sec:whatchanges}

\begin{figure}[t]
\centering
\includegraphics[width=\linewidth]{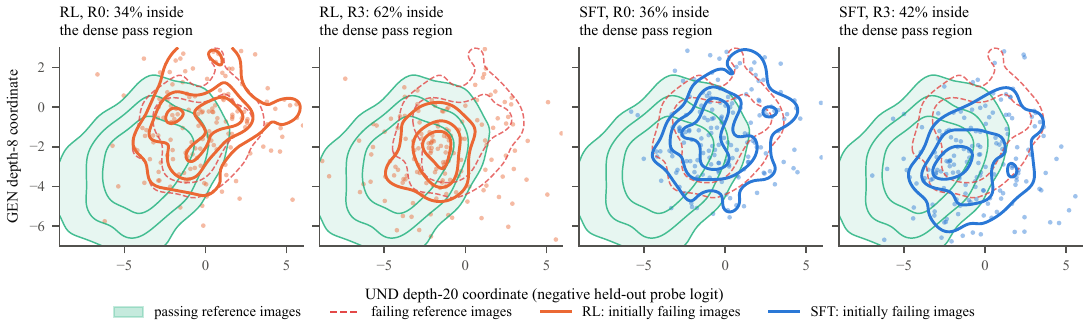}
\caption{\textbf{Image trajectories in the backbone's own correctness
readout.} Each point is one initially failing image, placed by two held-out
linear probes for ``passes the verifier'': the understanding stream at
depth 20 ($x$) and the generation stream at depth 8 ($y$), the depths where
each stream's probe AUC peaks. Green filled contours mark verifier-passing
reference images and red dashed contours verifier-failing ones; orange and
blue contours summarize the RL and SFT images. Read the panels in pairs,
from R0 to R3 for the same policy: RL (left pair) and SFT (right pair).
Percentages are the share inside the dense pass region
(Table~\ref{tab:distribution}).
}
\label{fig:distribution}
\label{fig:hypothesis_navigation}
\end{figure}

Replaying all 553 trajectories through the Base, SFT, and RL checkpoints
(Appendix~\ref{app:distribution}) gives one consistent picture. The
backbone's own readout of correctness is nearly unchanged by RL: a
held-out linear probe on the understanding stream reaches AUC 0.80--0.82
under all three checkpoints. What changes is where the images go. Among
initially failing images, the share inside the dense pass region of that
readout rises from 34\% to 62\% over three rounds under RL, against 36\%
to 42\% under SFT (Figure~\ref{fig:distribution}), and at a matched number of edits RL moves the image
further and more of its edits succeed. RL thus selects, among the revisions
the model can already produce, those that reach the correct region.

\section{Conclusion}
\label{sec:conclusion}

Reinforcement learning turns reflection in a unified model from an SFT cold
start into an effective repair mechanism. SFT and RL start from nearly
identical single-shot accuracy, so the +12-point GenEval gain and its
transfer to WISE, OneIG, and T2I-CompBench++ come from the reflection rounds.
RL leaves the model's correctness readout nearly unchanged and selects,
among the repairs the model can already produce, those that land: the
backbone already knows whether its image is correct, and RL teaches it to
act on that knowledge.

\bibliography{references}
\bibliographystyle{plainnat}

\clearpage
\appendix
\section*{Appendix}
\vspace{-0.2cm}
{
\small
\setlength{\parskip}{1pt}
\setlength{\baselineskip}{1.18\baselineskip}
\startcontents[appendices]
\printcontents[appendices]{l}{1}{\setcounter{tocdepth}{2}}
}
\clearpage
\FloatBarrier
\section{Implementation and Evaluation Details}
\label{app:protocol}

\paragraph{Training.}
The primary experiments use the BAGEL reflection-SFT checkpoint
(\S\ref{sec:sft}) as the RL parent.
RL samples from a $3{,}000$-prompt pool spanning six GenEval families;
a separate $270$-prompt held-out set is UID-disjoint from training.
Each update draws two independent roots with $K\!=\!16$ siblings per root.
Training uses 20 denoising steps and a two-transition flow window;
the final checkpoint is at $1{,}000$ updates.
RL runs on two nodes with eight NVIDIA H100 80GB GPUs each (16 GPUs,
hybrid-sharded FSDP); the $1{,}000$ updates take about 33 hours of update
time (median 118\,s per update).

\paragraph{Evaluation.}
All arms are evaluated with 50 denoising steps, $512\times512$ images,
and at most three native repairs.
BAGEL generates natively at $1024\times1024$; we set the generation
resolution to $512\times512$ for training and for every evaluated arm,
including Base. Because RL renders complete trajectories (up to four images
per rollout, 32 rollouts per update), the lower resolution keeps
whole-trajectory sampling tractable.
The same $1{,}000$-update checkpoint is shared across all benchmarks
rather than selecting a different checkpoint per test.
We evaluate one returned image per prompt per arm.

\paragraph{Benchmarks.}
\emph{GenEval} (553 prompts): unweighted macro accuracy over six
compositional families.
\emph{WISE} (1,000 prompts): weighted aggregate with a GPT-4o judge.
\emph{OneIG-Bench} (695 alignment prompts out of 1,120): question-dependent
alignment, not an overall omni-dimensional score.
\emph{T2I-CompBench++} (2,400 prompts, 300 per category): eight-category
mean with category-specific scorers.
Cross-benchmark numbers use the $0$--$100$ scale; native $0$--$1$ tables
are provided for GenEval.
Our controlled evaluations use instruction-voice prompts and differ from
native leaderboard protocols; we separate controlled comparisons from
published reference scores.

\begin{table}[ht]
\centering
\caption{Primary configuration and evaluation coverage.}
\small
\begin{tabular}{ll}
\toprule
Setting & Value \\
\midrule
Backbone & BAGEL, 28-layer mixture-of-transformers (MoT) \\
Initialization & Reflection SFT (one epoch) \\
Reported RL checkpoint & 1,000 committed updates, direct full weights \\
Direct T2I-RL control & Official Base, 1,000 RL updates, no repairs \\
Train/dev prompt pools & 3,000 / 270 \\
Roots per update / siblings per root & 2 / 16 \\
Training / inference denoising steps & 20 / 50 \\
Selected flow window & 2 contiguous transitions \\
Primary inference repair cap & 3 \\
Training / evaluation resolution & $512\times512$ (BAGEL native: $1024\times1024$) \\
RL hardware & 2 nodes $\times$ 8 NVIDIA H100 80GB \\
Evaluated outputs per prompt and arm & 1 \\
GenEval / WISE prompt counts & 553 / 1,000 \\
OneIG generated / alignment-scored & 1,120 / 695 \\
T2I-CompBench++ categories / prompts & 8 / 2,400 \\
SFT effective-record target / exposures & 167,363 / 167,368 \\
SFT continuation learning rate & $2\times10^{-7}$, constant \\
SFT CE / flow-MSE coefficients & 1 / 1 \\
RL text / flow learning rate & $5\times10^{-6}$ / $5\times10^{-6}$ \\
RL text / flow ratio clip $\epsilon$ & 0.2 / 0.1 \\
RL text / flow KL coefficient $\eta$ & $10^{-4}$ / $10^{-4}$ (frozen SFT reference) \\
KL estimator (text / flow) & per-token $k_3$ / closed-form Gaussian SDE transition \\
RL optimizer / weight decay & AdamW $(0.9,0.999)$ / $10^{-4}$ \\
RL gradient-norm clip & 1.0 per channel \\
Flow SDE noise level / text temperature & 1.0 / 0.9 \\
\bottomrule
\end{tabular}
\end{table}

\begin{figure}[ht]
\centering
\includegraphics[width=\linewidth]{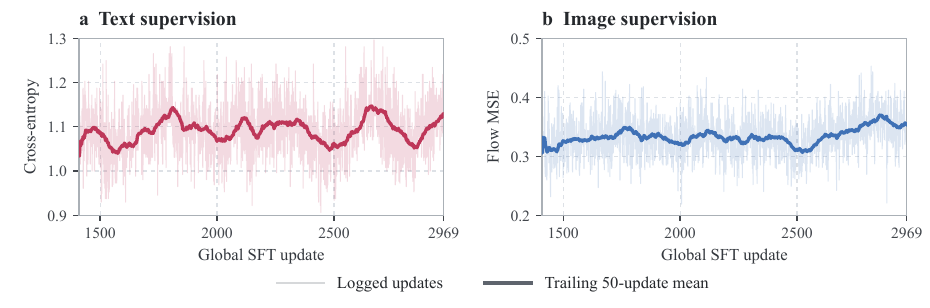}
\caption{\textbf{The supervised parent, recorded continuation phase.}
Text cross-entropy and image flow-MSE from global SFT updates 1410--2969.
Faint lines show logged update values and solid lines show trailing
50-update means.}
\label{fig:sft}
\end{figure}

\paragraph{Supervised parent provenance.}
The retained SFT parent completes an effective sample-exposure target of
167,363 records, reaching 167,368 at the final batch.
Global updates 1,410--2,969 run at a constant learning rate of $2\times10^{-7}$.
The mixture weights for controller, image-transition, verifier-state,
penultimate verifier-state, and base-generation anchor groups are
$4:4:4:1:1$. Figure~\ref{fig:sft} plots this recorded continuation phase.

\paragraph{Frozen and trainable modules.}
The RL implementation first freezes the model, then opens both expert
branches in all 28 MoT decoder layers and the flow roots
\texttt{time\_embedder}, \texttt{vae2llm}, \texttt{llm2vae}, and
\texttt{latent\_pos\_embed}. The remaining modules, including the vocabulary
head and the final normalization outside the decoder layers, remain frozen.
The two channel optimizers operate on explicit, disjoint parameter groups.
The detached initial rendering is not assigned an RL loss, although it can
change across updates because its generation parameters are also used for
repair.

\paragraph{Image-only versus protocol-gated accuracy.}
The image score is computed from the retained per-image verifier verdict
for the final returned image. We do not multiply it by a parse-validity
indicator. A malformed reflection can therefore leave a correct image
with a valid image score, while separately lowering protocol validity.
This separation is applied to SFT and RL alike. GenEval macro averages
weight the six families equally; prompt-micro averages instead weight all
553 prompts equally. Their distinct labels are preserved throughout.

\paragraph{Native and external-agent controls.}
The untuned external agent reviews the original request and current image
and always executes three corrections with the native BAGEL ODE sampler.
Its reviews are greedy and capped at 512 tokens. SFT and \method{} use the
same native reflection prompt, temperature 0.5 for controller sampling,
and a maximum of 512 controller tokens per turn. They stop on native
\textsc{done}, invalid termination, or the repair cap.

\paragraph{Benchmark-specific interpretation.}
WISE reports its weighted group aggregate, not a pooled accuracy.
OneIG reports question-dependency alignment on its 695 eligible prompts,
using the official Qwen2.5-VL-7B question-answering scorer.
T2I-CompBench++ uses one image per prompt and the prescribed category scoring
formulas. Its 3D-spatial category concerns relationships depicted in 2D
images, not generated 3D representations. Published native benchmark scores
are contextual references, not substitutes for matched re-evaluation.

\begin{figure}[ht]
\centering
\includegraphics[width=\linewidth]{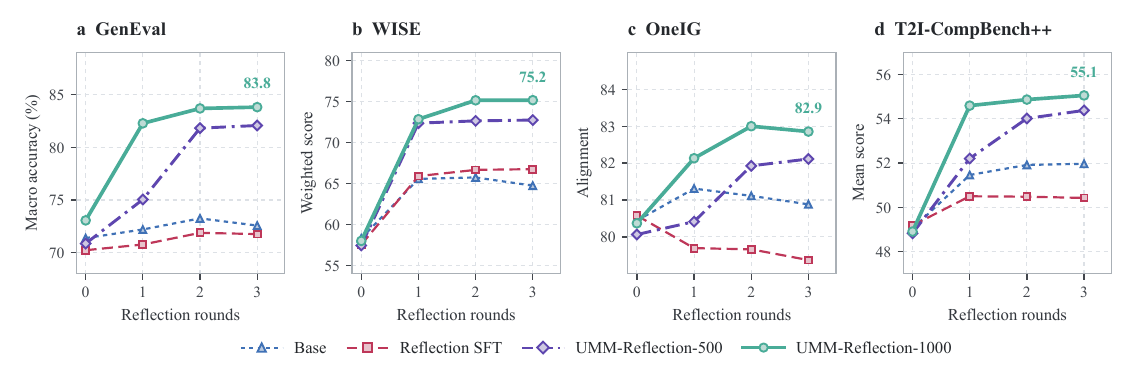}
\caption{\textbf{Multi-round test-time scaling on four benchmarks.}
Scores ($0$--$100$) versus reflection rounds.
Blue dotted lines denote Base, pink dashed lines reflection SFT,
purple dash-dotted lines \method{}-500, and green solid lines \method{}-1000.}
\label{fig:tts4}
\end{figure}

\paragraph{Test-time scaling evaluation.}
Figures~\ref{fig:tts} and~\ref{fig:tts4} use the retained images at each reflection round;
trajectories that stop early keep their last image.
WISE curves use a common GPT-4o re-evaluation of Base, SFT, and the
1000-step model. The main table reports the original WISE
evaluation.

\paragraph{RL prompt pool.}
The 3{,}000 RL prompts are drawn from the GenEval-style training prompts
released with Flow-GRPO \citep{flowgrpo2025}, restricted to the six GenEval
families and deduplicated so that no prompt repeats during training. Family
proportions follow Flow-GRPO's own mixture, with a fixed quota for single
object (position 1{,}305, counting 702, color attribution 559, colors 187,
two objects 187, single object 60), and
every prompt is rendered in the same instruction voice used at evaluation
(``Create an image with \dots''). No RL prompt reuses an official GenEval
evaluation prompt. The 270-prompt development set is disjoint from the
training pool. The pool is released with the code.

\paragraph{Representation probes.}
Correctness probes (Appendix~\ref{app:distribution}) are logistic
regressions (standardization, PCA to 256 dimensions fitted on the training
folds, $C=0.05$) on backbone features at depths $0,4,\dots,28$. They are
evaluated by five-fold cross-validation grouped by prompt, so images of the
same trajectory never appear in both training and test folds. AUC intervals
use 1{,}000 family-stratified prompt-bootstrap replicates.

\subsection{Direct-generation RL control}
\label{app:directrl}

The direct T2I-RL control initializes from official BAGEL Base and
optimizes direct T2I generation with Flow-GRPO, without reflection SFT or a
controller objective. It uses the same six-family training pool as
\method{}, $K=16$, 20 training denoising steps, two selected flow
transitions, and the original Base as the KL reference. The reported
checkpoint is at 1,000 updates.

Evaluation loads the direct T2I-RL language-model weights over Base
auxiliary modules and reuses the retained Base initial-generation path at
$512\times512$ with 50 denoising steps. There is no self-CoT, verifier
query, candidate selection, or repair at inference. GenEval preserves the
retained two-prompt seed batches; external benchmarks preserve one-prompt
calls and original indices/seeds. Coverage is complete for all 553 GenEval
prompts, 1,000 WISE prompts, 695 eligible OneIG alignment prompts out of
1,120 generated prompts, and 2,400 T2I-CompBench++ prompts.

\begin{figure}[ht]
\centering
\includegraphics[width=\linewidth]{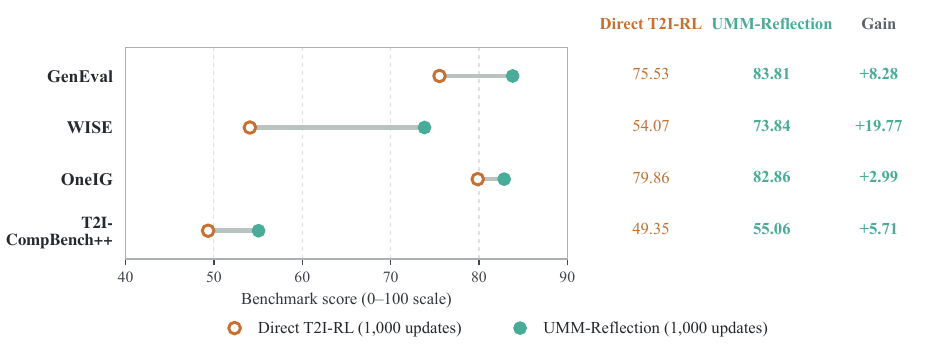}
\caption{\textbf{Completed direct-T2I RL control.}
Hollow orange points denote direct T2I-RL at 1,000 updates; green points denote
\method{} at 1,000 updates. Gain is the within-benchmark score difference.
Each benchmark keeps its own metric on a $0$--$100$ scale; scores are not
averaged across benchmarks. The systems differ in initialization and
training/inference compute (Appendix~\ref{app:protocol}).}
\label{fig:directrl}
\end{figure}

\subsection{Understanding-based selection protocol}
\label{app:undbon}

The two Best-of-4 arms draw four images per prompt from the direct T2I-RL
renderer over the full 553-prompt GenEval set, at $512\times512$ and 50
denoising steps. The first draw is the retained T2I-RL evaluation image; the
other three reuse its seed offset by multiples of $10^6$. The candidate order is shuffled
with seed $20260909+\text{prompt index}$ and is identical for both selectors.
Each makes one greedy multi-image call with \texttt{think=False} and a
256-token output limit. Base UND uses original Base; \method{} UND uses the \method{} checkpoint. ``UND'' identifies the understanding inference path, not a separately
trained classifier head.

The fixed instruction asks the model to prioritize requested objects,
counts, attributes and spatial relations over aesthetics, select the closest
visible match, and output \texttt{BEST: A/B/C/D} followed by a short explanation.
The complete selection prompt will be released with the code. Neither model receives the task family,
detector metadata, candidate scores, or selected-image verdict before choosing.
Each arm has one invalid-format response out of 553; the preregistered fallback
selects the first presented image, with no discarded prompts.
Image scoring follows the saved choices.

Base UND and \method{} UND select 437 and 440 correct images respectively
(prompt-micro counts, distinct from the macro scores).
Their A/B/C/D choice counts are 225/26/240/62 and 245/32/219/57.

\FloatBarrier
\section{Reflection Format and SFT Details}
\label{app:sftdetails}

Each reflection is exposed through six tagged fields: \texttt{[CURRENT\_ROUND]},
\texttt{[SOURCE\_IMAGE]}, \texttt{[SCORE]}, \texttt{[THINKING]},
\texttt{[ACTION]}, and \texttt{[EDIT]}, making the model's reasoning
inspectable.
The \texttt{[SCORE]} is a model-generated self-assessment, not an
external reward: neither verifier scores nor verifier labels enter the
policy observation at training or inference.

SFT teaches the interleaved protocol using the trajectories from
\S\ref{sec:data}.
Each trajectory is decomposed into $167{,}363$ training rows spanning
three complementary views: \emph{controller} rows supervise the full
reflection text (autoregressive cross-entropy, no image loss),
\emph{transition} rows supervise each individual edit step (the
edited image receives flow-matching loss conditioned on the edit
instruction), and \emph{verifier} rows present a partial trajectory
and supervise only the next reflection.
In notation,
\[
\mathcal{L}_{\text{SFT}}
  = \mathcal{L}_{\text{AR}}
  + \lambda_{\text{img}}\,\mathcal{L}_{\text{FM}}.
\]
Both understanding and generation expert branches across all 28
decoder layers are trainable; the visual encoders (ViT, VAE),
text embeddings, and vocabulary projection are frozen.
Training starts from the base BAGEL checkpoint and runs for one
full epoch.

\FloatBarrier
\section{Graded Reward}
\label{sec:graded}

Standard GenEval scoring is binary: an image either satisfies all
constraints or it does not.
Under binary scoring, five of the six GenEval families produce
$\Delta_t = 0$ whenever an edit improves some constraints but not all,
because $q$ jumps only at the boundary between full failure and full
success.
This renders all progress terms in Eq.~\ref{eq:reward} ineffective
for the majority of training prompts.

We therefore grade only the failure region, using nothing but what the
frozen verifier returns at its official thresholds ($0.3$ for object
detection, $0.9$ for counting); no sub-threshold confidence is read. For
every family,
\[
q=\begin{cases}1 & \text{official verdict passes},\\
\min\!\bigl(\tfrac12 f,\ 0.5^{-}\bigr) & \text{otherwise},\end{cases}
\]
where $f\in[0,1]$ is a family-specific satisfaction fraction and $0.5^{-}$ is
the largest double below $0.5$, so partial credit never reaches the exact
band. Let $\pi(o)\in\{0,1\}$ indicate that the verifier keeps a detection of
class $o$; missing detections and boxes contribute zero.
\begin{itemize}
\item \emph{single object}: $f=\pi(o)$.
\item \emph{two objects}: $f=\tfrac12\bigl(\pi(o_1)+\pi(o_2)\bigr)$.
\item \emph{colors}: $f=\pi(o)\bigl(\tfrac12+\tfrac12 c\bigr)$, where $c$ is the
CLIP confidence of the predicted color if it is the requested one and $0$
otherwise.
\item \emph{color attribution}: $f=\tfrac12(r_1+r_2)$, with $r_k=1$ if
attribute $k$ passes and otherwise $r_k=\pi(o_k)\bigl(\tfrac12+\tfrac12
c_k\bigr)$, $c_k=\tfrac12\bigl(c_k^{\text{CLIP}}+\operatorname{clip}(1+b_k,0,1)\bigr)$;
$b_k$ is the BLIP margin (yes-probability of the requested phrase minus that
of the color-swapped phrase), whose official cut is $0$.
\item \emph{position}: $f=0.6\cdot\tfrac12(\pi_s+\pi_o)+0.4\,\min(\pi_s,\pi_o)\,m$,
with $m=\operatorname{clip}(d/0.5,0,1)$, where $d$ is the component of the
official threshold-shrunk, normalized center offset along the requested
relation (the official pass cut is $0.5$; $m=0$ if either box is missing).
\item \emph{counting}: $q=1$ if $\hat n=n$ and
$q=\tfrac12\max\bigl(0,1-|\hat n-n|/n\bigr)$ otherwise, where $\hat n$ is the
number of detections at the counting threshold.
\end{itemize}

\paragraph{Fail-closed rule.}
Every scored image is checked against its official verdict: a passing image
must receive $q=1$ and a failing image $q<0.5$. A violation raises an error in
the scoring service and the request returns no reward, so grading can never
change a pass/fail decision, and accuracy computed from $q$ equals the binary
GenEval score. On 200 image--family pairs scored with the frozen backends, no
violation occurred and the share of images with a nonzero score rose from
21\% to 100\%.

Malformed trajectories receive a bounded negative correction
$A_i \leftarrow \max\bigl(\min(A_i, 0) - 0.5,\,-1\bigr)$, preserving a learning signal
for protocol violations.

\FloatBarrier
\section{Training Dynamics}
\label{app:training}

\begin{figure}[ht]
\centering
\includegraphics[width=\linewidth]{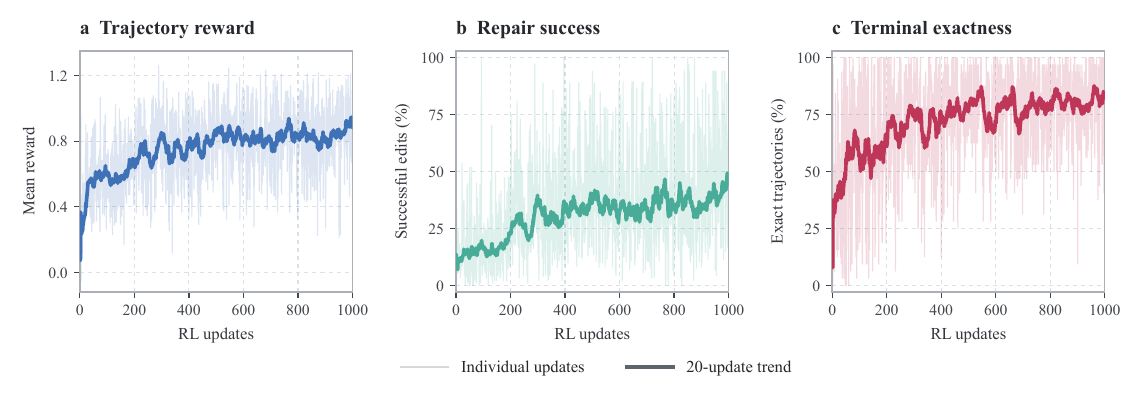}
\caption{\textbf{Learning dynamics of the 1,000-update RL run.}
(a) Mean trajectory reward. (b) Successful repairs per edit from an
incorrect state. (c) Fraction of trajectories reaching terminal exactness
under the training verifier. Faint lines are individual updates; solid
lines are trailing 20-update trends, with rates pooling numerators and
denominators over the window.}
\label{fig:training}
\end{figure}

\FloatBarrier
\section{Test-Time Scaling on External Benchmarks}
\label{app:tts}

\FloatBarrier
\section{Distributional Analysis}
\label{sec:distribution}
\label{sec:hypotheses}
\label{app:distribution}

The findings below support one reading: RL does not give the model a new
way to see or judge its images; among the revisions the backbone can already
produce, it learns to select the ones that land, a pattern also reported for
RL with verifiable rewards in language models \citep{yue2025rlbeyond}.
We replayed all 553 archived RL trajectories, together with the matching
SFT trajectories, through the Base, SFT, and RL checkpoints and read out
both the generation branch (VAE latents) and the understanding branch
(ViT tokens) at every decoder depth (protocol in
Appendix~\ref{app:protocol}). Two findings organize what changed.

\begin{table}[ht]
\centering
\caption{\textbf{Distributional statistics.} One trajectory per prompt. Edit-matched rows use prompts where SFT made three edits
(RL always makes three) or at least one edit. Intervals are 95\%
family-stratified prompt bootstrap intervals, except the pass-region change,
which resamples images (5{,}000 replicates). Pass-region rows use
initially failing images.}
\label{tab:distribution}
\small
\setlength{\tabcolsep}{4pt}
\begin{adjustbox}{max width=\linewidth}
\begin{tabular}{@{}llrrr@{}}
\toprule
Measure & Unit ($N$) & SFT & RL & RL $-$ SFT [95\% CI] \\
\midrule
DINO distance, first to last image & prompts (553) & 0.09 & 0.29 & -- \\
\quad same, three edits & prompts (153) & 0.11 & 0.28 & 0.17 [0.13, 0.21] \\
Exact gain R0$\to$R3 (points), three edits & prompts (153) & $-3.3$ & $+16.3$ & $+19.6$ [11.1, 28.1] \\
Exact gain R0$\to$R3 (points), $\geq$1 edit & prompts (456) & $+0.9$ & $+10.3$ & -- \\
In dense pass region, R0$\to$R3 (\%) & images (170 / 154) & 36.5$\to$42.4 & 34.4$\to$61.7 & $+21.4$ [9.4, 33.3] \\
UND probe AUC, depth 20 (Base 0.804) & held-out images & 0.807 & 0.815 & -- \\
Stream coupling, GEN / UND (null 0.04) & image pairs & 0.56 / 0.45 & 0.57 / 0.46 & Holm $p=0.012$ vs.\ null \\
\bottomrule
\end{tabular}
\end{adjustbox}
\end{table}

\paragraph{RL edits move the image further, at matched edit count.}
Over all 553 prompts, the DINO distance from the first to the last image
is 0.09 for SFT and 0.29 for RL. This is not because RL edits more often.
On the 153 prompts where both policies used exactly three edits, the
distance is 0.11 versus 0.28 (paired difference 0.17, 95\% CI
[0.13,\,0.21]), and on the same prompts the exact-match gain from R0 to R3
is $-3.3$ points for SFT and $+16.3$ for RL (paired difference $+19.6$
[11.1,\,28.1]). On the 456 prompts where SFT made at least one edit, the
gains are $+0.9$ and $+10.3$. At equal edit budget, RL makes larger
changes and they land.

\paragraph{RL moves failing images into the passing region of a readout the backbone already has.}
A linear probe for ``passes the verifier'' on the understanding stream
peaks at decoder depth 20 with held-out AUC 0.804 under Base, 0.807 under
SFT, and 0.815 under RL, whereas frozen DINO and CLIP features reach only
0.55. The backbone can already tell whether its image is right, and RL
barely changes that readout. What RL changes is where the images go
(Figure~\ref{fig:hypothesis_navigation}): among initially failing images,
the share inside the dense pass region rises from 34\% at R0 to 62\% at
R3 under RL, versus 36\% to 42\% under SFT, and the probe distance of
failing RL images to the passing side falls monotonically over the rounds
under all three checkpoints. Read against SFT, the figure shows where the
gain comes from. SFT's revisions spread broadly over the readout plane, and
only a small part of that distribution reaches the dense pass region. RL
does not open a new region: its R3 images concentrate inside the same pass
region that part of SFT's distribution already reaches. RL extracts the
correct slice of the broad distribution that SFT learned and shifts the
policy's trajectories toward it; the rollout statistics in
Appendix~\ref{app:passk} show the same pattern directly. The same holds for the coupling between the
two streams: changing the image changes the representation of a fixed
request, and the normalized alignment between the two displacements is
0.58/0.45 (GEN/UND) for Base, 0.56/0.45 for SFT, and 0.57/0.46 for RL,
each far above the permutation null of 0.04 (Holm $p=0.012$). RL does not
build a new channel between seeing and describing; it steers images
through one the unified backbone already has, which is why 1,000 updates
on a 3,000-prompt pool suffice.

\FloatBarrier
\section{Visual Pathway Stability}
\label{sec:attention}

\begin{figure}[ht]
\centering
\includegraphics[width=\linewidth]{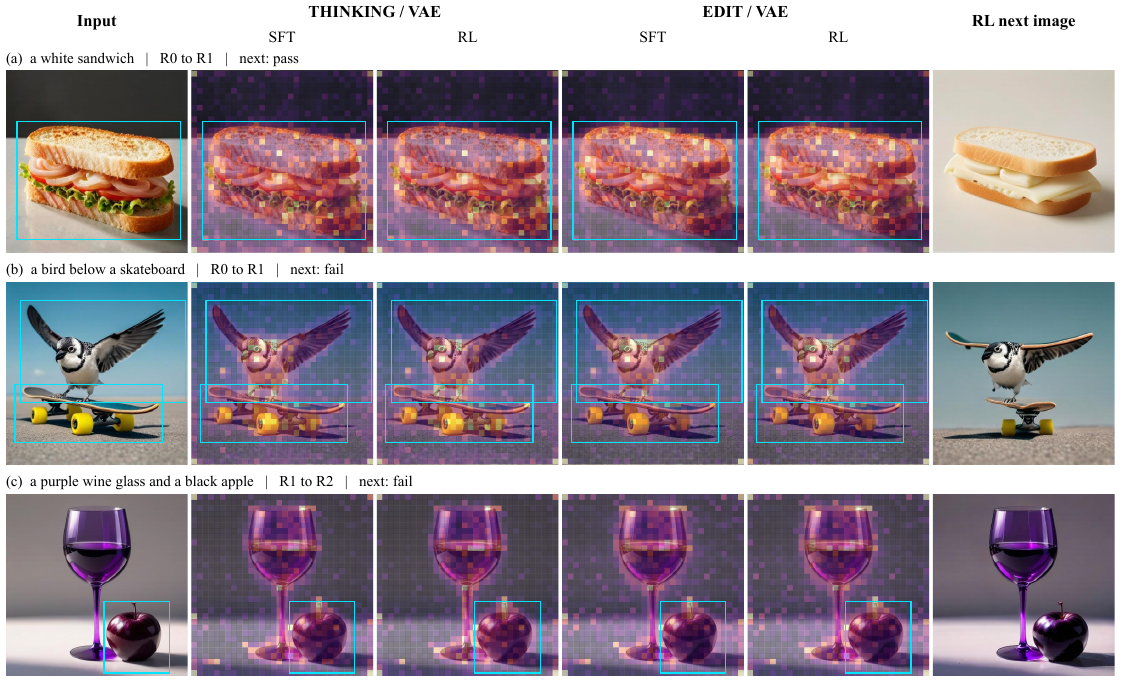}
\caption{\textbf{Attention over VAE image tokens is preserved after RL.}
Three cases show paired SFT/RL attention of \texttt{THINKING} and
\texttt{EDIT} tokens onto $32\times32$ VAE keys, averaged over 28
layers. Maps use a shared 99th-percentile cap; cyan boxes are
pre-annotated error regions.}
\label{fig:attention}
\end{figure}

Consistent with the distributional findings in Appendix~\ref{app:distribution}, the visual pathway itself is
largely unchanged by RL. On ten replayed trajectories (30 paired
rounds), SFT and RL attention maps over VAE keys correlate at a median
of 0.98; over ViT keys the correlation exceeds 0.99
(Figure~\ref{fig:attention}). The cross-modal coupling that connects
image states to text representations is already present in the Base
model, and RL preserves it. We observe
only limited changes in the visual pathway, which is consistent with the
learned change residing mainly in what the model writes from the same
visual input.

\FloatBarrier
\section{Cross-Round Attention to Earlier Images}
\label{app:crossround}

Each reflection round keeps all earlier images in context. To test whether
later rounds use them, we measure, on the ten replayed trajectories of
Appendix~\ref{sec:attention}, how the attention that \texttt{THINKING} and
\texttt{EDIT} tokens assign to image keys (VAE and ViT) is split across the
images in context. Table~\ref{tab:crossround} reports the third reflection
round, which sees the initial image $x_0$ and two revisions. About 45--48\% of
the image attention goes to the current image, but $x_0$ and the first
revision each keep 22--31\%, and every earlier image receives at least 19\%
in every case. Restricted to VAE keys, the three images are attended almost
equally (29--37\%). SFT and RL split attention the same way: the model reads
its whole visual history, and RL does not change this. 

\begin{table}[ht]
\centering
\caption{\textbf{Share of image attention per image at the third reflection
round.} Mean over ten trajectories (minimum in parentheses); shares sum to 1
across the three images. Image attention is 10--13\% of all attention.}
\label{tab:crossround}
\small
\begin{tabular}{@{}llccc@{}}
\toprule
Policy & Tokens & $x_0$ & Revision 1 & Current image \\
\midrule
SFT & \texttt{THINKING} & 0.23 (0.20) & 0.30 (0.26) & 0.48 (0.41) \\
SFT & \texttt{EDIT} & 0.23 (0.19) & 0.31 (0.27) & 0.46 (0.41) \\
RL & \texttt{THINKING} & 0.26 (0.23) & 0.29 (0.25) & 0.45 (0.39) \\
RL & \texttt{EDIT} & 0.23 (0.19) & 0.31 (0.27) & 0.46 (0.40) \\
\midrule
RL, VAE keys only & \texttt{THINKING} & 0.32 (0.27) & 0.34 (0.28) & 0.34 (0.26) \\
RL, VAE keys only & \texttt{EDIT} & 0.32 (0.27) & 0.35 (0.29) & 0.33 (0.26) \\
\bottomrule
\end{tabular}
\end{table}

\FloatBarrier
\section{Correct Repairs Already in the SFT Rollout Distribution}
\label{app:passk}

Each RL update samples $K=16$ complete trajectories from one initial image,
and the training log records whether each trajectory ends verifier-correct.
Table~\ref{tab:passk} uses the roots whose initial image is incorrect and
reports the per-trajectory success rate (pass@1) and the share of roots with
at least one correct trajectory among the 16 (pass@16), by training window.
At the start of RL, where the policy is essentially the SFT model, a correct
repair already exists among the 16 rollouts for 78\% of roots, although a
single rollout succeeds only 23\% of the time. Over training, pass@1 rises to
70\% while pass@16 rises only to 97\%: RL concentrates the policy on repairs
that the SFT distribution already contains.

\begin{table}[ht]
\centering
\caption{\textbf{Sibling success on incorrect initial images during RL.}
Roots with an incorrect initial image; 16 sibling trajectories per root.}
\label{tab:passk}
\small
\begin{tabular}{@{}lrrr@{}}
\toprule
RL updates & Roots & pass@1 (\%) & pass@16 (\%) \\
\midrule
1--50 & 67 & 22.7 & 77.6 \\
51--100 & 54 & 31.4 & 77.8 \\
101--200 & 114 & 31.5 & 83.3 \\
201--500 & 367 & 60.5 & 94.6 \\
501--1{,}000 & 611 & 70.1 & 97.1 \\
\bottomrule
\end{tabular}
\end{table}

\FloatBarrier
\section{Swapping the Editing Instruction on Fixed Images}
\label{app:swap}

To test which channel carries the repair, we fix the image, the RL
renderer, and the sampling noise, and change only the editing instruction
for one image update. On 154 initially failing states with two paired noise
seeds each, the original request repairs 20.5\%, the SFT policy's
instruction 21.4\%, and the RL policy's instruction 48.4\%; a rule-written
instruction that spells out every GenEval constraint reaches 27.6\%.
The paired RL-minus-request difference is $+27.9$ points (95\% state
bootstrap interval $[20.8, 34.7]$), and RL-minus-rule is $+20.8$
$[13.6, 27.9]$, concentrated in counting, position, and two-object prompts.
With the renderer and noise held fixed, the instruction the RL policy writes
is what turns a failing image into a correct one, and it is more executable
than an exhaustive rule-based specification.

\FloatBarrier
\section{Paired Final-Model Results}
\label{app:paired}

\begin{table}[ht]
\centering
\caption{\textbf{Stage-wise GenEval analysis.} Initial and final are
prompt-micro image accuracy over 553 prompts, not six-family macro scores.
Repair is conditional on an initially incorrect image; damage is
conditional on an initially correct image. Protocol validity is measured
separately. All entries are percentages.}
\label{tab:stages}
\small
\setlength{\tabcolsep}{5pt}
\begin{adjustbox}{max width=\linewidth}
\begin{tabular}{lrrrrr}
\toprule
Model & Initial & Final & Repair & Damage & Protocol \\
\midrule
Reflection SFT & 69.26 & 71.07 & 20.59 & 6.53 & 95.48 \\
\method{} (1,000 updates) & 72.15 & 83.91 & 64.94 & 8.77 & 100.00 \\
\bottomrule
\end{tabular}

\end{adjustbox}
\end{table}

\begin{table}[ht]
\centering
\caption{Prompt-paired \method{}-1000 versus reflection SFT on GenEval.
Wins and losses count discordant image-only verdicts. The exact two-sided
binomial test on discordant pairs is the exact McNemar test.}
\small
\begin{tabular}{lrrrr}
\toprule
Comparison with SFT & Wins & Losses & Net & Exact $p$ \\
\midrule
Initial & 48 & 32 & +16 & 0.093 \\
Final & 109 & 38 & +71 & $<0.001$ \\
\bottomrule
\end{tabular}

\end{table}

All tables in this section compare the retained SFT evaluation with the
final full-weight RL model on the same 553 prompts. Earlier decoder-only
checkpoint evaluations are not substituted for this final-model evaluation.

\section{Category-Level Results}
\label{app:categories}

\begin{table}[ht]
\centering
\caption{All six GenEval families, using image-only verdicts. Values are
percentages; the macro mean appears in Table~\ref{tab:geneval}.}
\small
\begin{tabular}{lrrrr}
\toprule
Family & $n$ & SFT final & RL initial & RL final \\
\midrule
Single object & 80 & 100.00 & 100.00 & 95.00 \\
Two objects & 99 & 87.88 & 83.84 & 95.96 \\
Counting & 80 & 57.50 & 67.50 & 67.50 \\
Colors & 94 & 87.23 & 84.04 & 90.43 \\
Position & 100 & 47.00 & 55.00 & 89.00 \\
Color binding & 100 & 51.00 & 48.00 & 65.00 \\
\bottomrule
\end{tabular}

\end{table}

\begin{figure}[ht]
\centering
\includegraphics[width=\linewidth]{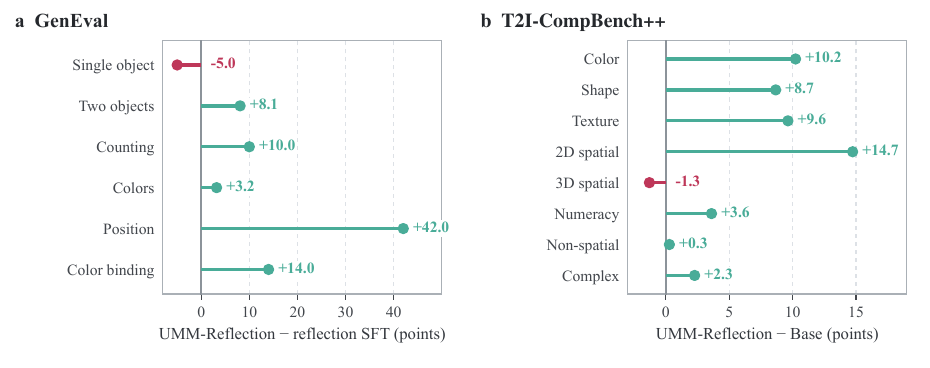}
\caption{Category-level changes, with every category retained.
Positive changes are green and negative changes are red. This view
complements the overall gains without assuming uniform improvement.}
\end{figure}

\FloatBarrier
\section{Detailed External Benchmark Comparisons}
\label{app:external}

These detailed tables use native $0$--$1$ scores. GenEval and original WISE
sources generally report two decimal places, OneIG three, and T2I-CompBench++
four; we preserve these precisions and print local measurements to four
places. T2I-CompBench++ reference means are the eight-category mean of the
published category scores. The overview, ablation, stage statistics, and
plots retain their stated $0$--$100$ scales.

GenEval row citations distinguish model-author results, the original
benchmark-author results, and secondary reports marked $\dagger$.
Resolution notes for older reference models additionally follow
\citet{tokenflow2024}, Table 4; undocumented settings are marked
\textit{n/r}, rather than assumed to be 512.
Other native-resolution reference models are not protocol-matched controls.
Each reference row reports the source table of the cited paper.

\begin{table}[ht]
\centering
\caption{\textbf{WISE: six world-knowledge categories.}
Native $0$--$1$ results from the official original-WISE legacy leaderboard
\citep{wise2025}, retaining its two-decimal reporting precision.
The local block uses 1,000 original prompts and the same frozen judge setup.
Overall is the original weighted aggregate. These are not
WISE\_Verified scores: that revision changes prompts, judge, and scoring.
Bold denotes the best completed local result.}
\label{tab:wise}
\fontsize{9}{10.5}\selectfont
\setlength{\tabcolsep}{4pt}
\begin{adjustbox}{max width=\linewidth}
\begin{tabular}{@{}lrrrrrrr@{}}
\toprule
Model & Cultural & Time & Space & Biology & Physics & Chem. & Overall \\
\midrule
\multicolumn{8}{l}{\textit{Official original-WISE legacy leaderboard: \citet{wise2025}}} \\
SD1.5 & 0.34 & 0.35 & 0.32 & 0.28 & 0.29 & 0.21 & 0.32 \\
SDXL (base 0.9) & 0.43 & 0.48 & 0.47 & 0.44 & 0.45 & 0.27 & 0.43 \\
SD3.5-Large & 0.44 & 0.50 & 0.58 & 0.44 & 0.52 & 0.31 & 0.46 \\
PixArt-$\alpha$ & 0.45 & 0.50 & 0.48 & 0.49 & 0.56 & 0.34 & 0.47 \\
Playground v2.5 & 0.49 & 0.58 & 0.55 & 0.43 & 0.48 & 0.33 & 0.49 \\
FLUX.1-dev & 0.48 & 0.58 & 0.62 & 0.42 & 0.51 & 0.35 & 0.50 \\
Janus-1.3B & 0.16 & 0.26 & 0.35 & 0.28 & 0.30 & 0.14 & 0.23 \\
VILA-U-7B-256 & 0.26 & 0.33 & 0.37 & 0.35 & 0.39 & 0.23 & 0.31 \\
Show-o-512 & 0.28 & 0.40 & 0.48 & 0.30 & 0.46 & 0.30 & 0.35 \\
Janus-Pro-7B & 0.30 & 0.37 & 0.49 & 0.36 & 0.42 & 0.26 & 0.35 \\
Emu3 & 0.34 & 0.45 & 0.48 & 0.41 & 0.45 & 0.27 & 0.39 \\
MetaQuery-XL & 0.56 & 0.55 & 0.62 & 0.49 & 0.63 & 0.41 & 0.55 \\
\midrule
\multicolumn{8}{l}{\textit{Local evaluation: same prompts, resolution, and scorer}} \\
BAGEL-Base & 0.4841 & 0.5437 & 0.6774 & 0.5005 & 0.6445 & 0.4590 & 0.5515 \\
BAGEL-SFT & 0.5838 & 0.6254 & 0.7165 & 0.6120 & 0.6965 & 0.5380 & 0.6287 \\
BAGEL-Self-Agentic & 0.5921 & 0.5967 & 0.7410 & 0.5550 & 0.7010 & 0.4915 & 0.6129 \\
BAGEL-T2I-RL (1k) & 0.4829 & 0.5284 & 0.6711 & 0.4620 & 0.6595 & 0.4405 & 0.5407 \\
\rowcolor{bakerlrow}\method{} & \textbf{0.6835} & \textbf{0.7054} & \textbf{0.8226} & \textbf{0.7925} & \textbf{0.7985} & \textbf{0.6280} & \textbf{0.7384} \\
\bottomrule
\end{tabular}

\end{adjustbox}
\end{table}

\begin{table}[ht]
\centering
\caption{\textbf{T2I-CompBench++: all eight composition categories.}
Native $0$--$1$ scores retain four decimal places. The public block
transcribes the non-MLLM columns of
\citet{compbench2025}, Table XIII: BLIP-VQA for attributes, UniDet for
2D/3D spatial relations and numeracy, CLIP for non-spatial relations, and
3-in-1 for complex composition. Published evaluation uses ten images per
prompt; our local block uses one image for each of 2,400 prompts.
The eight-category mean is our local aggregate and is not supplied for
published rows. Bold denotes the best completed local result.}
\label{tab:compbench}
\fontsize{8}{9.5}\selectfont
\setlength{\tabcolsep}{3pt}
\begin{adjustbox}{max width=\linewidth}
\begin{tabular}{@{}lrrrrrrrrr@{}}
\toprule
Model & Color & Shape & Texture & 2D & 3D & Number & Non-sp. & Complex & Mean \\
\midrule
\multicolumn{10}{l}{\textit{Published ten-image protocol: \citet{compbench2025}, Table XIII}} \\
SD1.4 & 0.3765 & 0.3576 & 0.4156 & 0.1246 & 0.3030 & 0.4456 & 0.3079 & 0.3080 & 0.3298 \\
SD2 & 0.5065 & 0.4221 & 0.4922 & 0.1342 & 0.3230 & 0.4582 & 0.3127 & 0.3386 & 0.3734 \\
Composable + SD2 & 0.4063 & 0.3299 & 0.3645 & 0.0800 & 0.2847 & 0.4272 & 0.2980 & 0.2898 & 0.3100 \\
Structured + SD2 & 0.4990 & 0.4218 & 0.4900 & 0.1386 & 0.3224 & 0.4557 & 0.3111 & 0.3355 & 0.3718 \\
Attend-and-Excite + SD2 & 0.6400 & 0.4517 & 0.5963 & 0.1455 & 0.3222 & 0.4773 & 0.3109 & 0.3401 & 0.4105 \\
GORS-unbiased + SD2 & 0.6414 & 0.4546 & 0.6025 & 0.1725 & 0.3300 & 0.4849 & 0.3158 & 0.3470 & 0.4186 \\
GORS + SD2 & 0.6603 & 0.4785 & 0.6287 & 0.1815 & 0.3572 & 0.4830 & 0.3193 & 0.3328 & 0.4302 \\
SDXL & 0.5879 & 0.4687 & 0.5299 & 0.2133 & 0.3566 & 0.4991 & 0.3119 & 0.3237 & 0.4114 \\
PixArt-$\alpha$-ft & 0.6690 & 0.4927 & 0.6477 & 0.2064 & 0.3901 & 0.5032 & 0.3197 & 0.3433 & 0.4465 \\
DALL$\cdot$E 3 & 0.7785 & 0.6205 & 0.7036 & 0.2865 & 0.3744 & 0.5926 & 0.3003 & 0.3773 & 0.5042 \\
SD3 & 0.8132 & 0.5885 & 0.7334 & 0.3200 & 0.4084 & 0.6174 & 0.3140 & 0.3771 & 0.5215 \\
FLUX.1 & 0.7407 & 0.5718 & 0.6922 & 0.2863 & 0.3866 & 0.6185 & 0.3127 & 0.3703 & 0.4974 \\
\midrule
\multicolumn{10}{l}{\textit{Local evaluation: same prompts, resolution, and scorer}} \\
BAGEL-Base & 0.7437 & 0.5216 & 0.6658 & 0.3015 & 0.3913 & 0.6067 & 0.3079 & 0.3847 & 0.4904 \\
BAGEL-SFT & 0.7609 & 0.5725 & 0.7123 & 0.3014 & 0.3759 & 0.6182 & 0.3061 & 0.3870 & 0.5043 \\
BAGEL-Self-Agentic & 0.8010 & 0.5741 & 0.7203 & 0.3268 & \textbf{0.4090} & 0.6300 & 0.3059 & 0.3905 & 0.5197 \\
BAGEL-T2I-RL (1k) & 0.7402 & 0.5029 & 0.6684 & 0.3150 & 0.3931 & 0.6408 & \textbf{0.3126} & 0.3746 & 0.4935 \\
\rowcolor{bakerlrow}\method{} & \textbf{0.8460} & \textbf{0.6083} & \textbf{0.7621} & \textbf{0.4489} & 0.3784 & \textbf{0.6428} & 0.3108 & \textbf{0.4074} & \textbf{0.5506} \\
\bottomrule
\end{tabular}

\end{adjustbox}
\end{table}

\begin{table}[ht]
\centering
\caption{\textbf{OneIG-Bench alignment.} Native $0$--$1$ scores.
Reference rows are from \citet{oneig2025}, Table 2, at their reported
precision and resolution (Table 8). Local rows measure question-dependent
alignment on the 695 eligible prompts at $512^2$ with the same scorer.}
\label{tab:oneig}
\fontsize{9}{10.5}\selectfont
\setlength{\tabcolsep}{5pt}
\begin{adjustbox}{max width=\linewidth}
\begin{tabular}{@{}lrr@{}}
\toprule
Model & Res. & Alignment \\
\midrule
\multicolumn{3}{l}{\textit{Benchmark-author results: \citet{oneig2025}, Tables 2 and 8}} \\
Janus-Pro & 384 & 0.553\phantom{0} \\
BLIP3-o & 1024 & 0.711\phantom{0} \\
Show-o2-1.5B & 432 & 0.798\phantom{0} \\
Show-o2-7B & 432 & 0.817\phantom{0} \\
OmniGen2 & 1024 & 0.804\phantom{0} \\
SD1.5 & 512 & 0.565\phantom{0} \\
SDXL & 1024 & 0.688\phantom{0} \\
SD3.5-Large & 1024 & 0.809\phantom{0} \\
FLUX.1-dev & 1024 & 0.786\phantom{0} \\
CogView4 & 1024 & 0.786\phantom{0} \\
SANA-1.5 1.6B (PAG) & 1024 & 0.762\phantom{0} \\
SANA-1.5 4.8B (PAG) & 1024 & 0.765\phantom{0} \\
Lumina-Image 2.0 & 1024 & 0.819\phantom{0} \\
HiDream-I1-Full & 1024 & 0.829\phantom{0} \\
\midrule
\multicolumn{3}{l}{\textit{Local evaluation: same prompts, resolution, and scorer}} \\
BAGEL-Base & 512 & 0.8044 \\
BAGEL-SFT & 512 & 0.7937 \\
BAGEL-Self-Agentic & 512 & 0.8088 \\
BAGEL-T2I-RL (1k) & 512 & 0.7986 \\
\rowcolor{bakerlrow}\method{} & 512 & \textbf{0.8286} \\
\bottomrule
\end{tabular}

\end{adjustbox}
\end{table}

\begin{table}[ht]
\centering
\caption{\textbf{OneIG local alignment by prompt category.}
Scores use the native $0$--$1$ scale. Anime/stylization, general objects,
and portrait contain 245, 206, and 244
eligible prompts, respectively. Alignment is the prompt-weighted aggregate,
not the unweighted mean of these three columns. The Anime/style column
measures alignment on that prompt category, not the separate style metric
in Table~\ref{tab:oneig}.}
\label{tab:oneigcategories}
\small
\setlength{\tabcolsep}{4pt}
\begin{adjustbox}{max width=\linewidth}
\begin{tabular}{@{}lrrrr@{}}
\toprule
Model & Anime / style & General objects & Portrait & Alignment \\
\midrule
\multicolumn{5}{l}{\textit{Local evaluation: same prompts, resolution, and scorer}} \\
BAGEL-Base & 0.8389 & 0.7569 & 0.8099 & 0.8044 \\
BAGEL-SFT & 0.8273 & 0.7529 & 0.7945 & 0.7937 \\
BAGEL-Self-Agentic & 0.8436 & 0.7732 & 0.8040 & 0.8088 \\
BAGEL-T2I-RL (1k) & 0.8326 & 0.7408 & 0.8133 & 0.7986 \\
\rowcolor{bakerlrow}\method{} & \textbf{0.8618} & \textbf{0.8020} & \textbf{0.8176} & \textbf{0.8286} \\
\bottomrule
\end{tabular}

\end{adjustbox}
\end{table}

\FloatBarrier
\section{External Critic Baseline}
\label{app:gptcritic}

We also compare with an external pipeline in which GPT-5.5
(\texttt{gpt-5.5-2026-04-23}, medium effort) inspects each BAGEL-Base image
and issues either an edit instruction or \textsc{done}; Base executes up to
three edits. The critic sees only the request and the current image, never
a verifier verdict, and the protocol otherwise matches our evaluation
(same prompts, seeds, resolution, and 50 denoising steps).
Table~\ref{tab:gptcritic} reports the result. On GenEval the pipeline
reaches 79 and repairs 28\% of the 163 initially incorrect images
(45 of 163), against 84 and 65\% for \method{}; GPT-5.5 accepts 67 of the
163 incorrect initial images without a single edit. \method{} also leads on
WISE and T2I-CompBench++ and matches the pipeline on OneIG, with no external model
at inference.

\begin{table}[ht]
\centering
\caption{\textbf{External GPT-5.5 critic versus \method{}.} Native $0$--$1$
scale. WISE in this table is scored by a separate GPT-4o run
for all three rows. OneIG to three decimals: 0.825 (critic) versus 0.829 (\method{}).
Repair is the share of initially incorrect GenEval images that end correct.}
\label{tab:gptcritic}
\small
\setlength{\tabcolsep}{5pt}
\begin{tabular}{@{}lrrrrr@{}}
\toprule
Model & GenEval & WISE & OneIG & T2I-CompBench++ & Repair (\%) \\
\midrule
BAGEL-Base & 0.71 & 0.58 & 0.80 & 0.49 & -- \\
BAGEL + GPT-5.5 critic & 0.79 & 0.70 & 0.83 & 0.52 & 27.6 \\
\rowcolor{bakerlrow}\method{} & \textbf{0.84} & \textbf{0.75} & \textbf{0.83} & \textbf{0.55} & \textbf{64.9} \\
\bottomrule
\end{tabular}
\end{table}

\FloatBarrier
\section{Additional Qualitative Examples}
\label{app:qualitative}

\begin{figure}[ht]
\centering
\includegraphics[width=\linewidth]{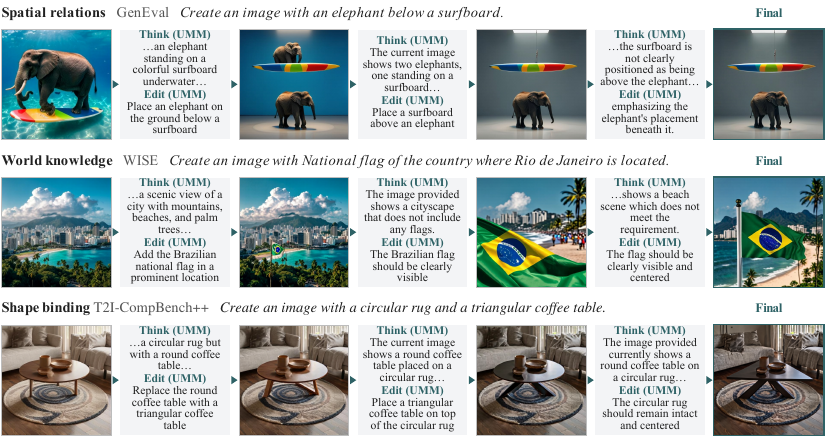}
\caption{\textbf{Reflection trajectories of \method{}.}
Examples from GenEval, WISE, and T2I-CompBench++ show how step-by-step reflection and revision help the model produce images that match the prompt. Each row reads left to right: the image, then the unified model's own reflection on it (\emph{Think}: what is wrong; \emph{Edit}: the instruction it issues), then the image it renders from that instruction. All text is verbatim model output. 
}
\label{fig:teaser}
\end{figure}

All images and text excerpts in this section are unedited model outputs.

\begin{figure}[p]
\centering
\includegraphics[width=\linewidth]{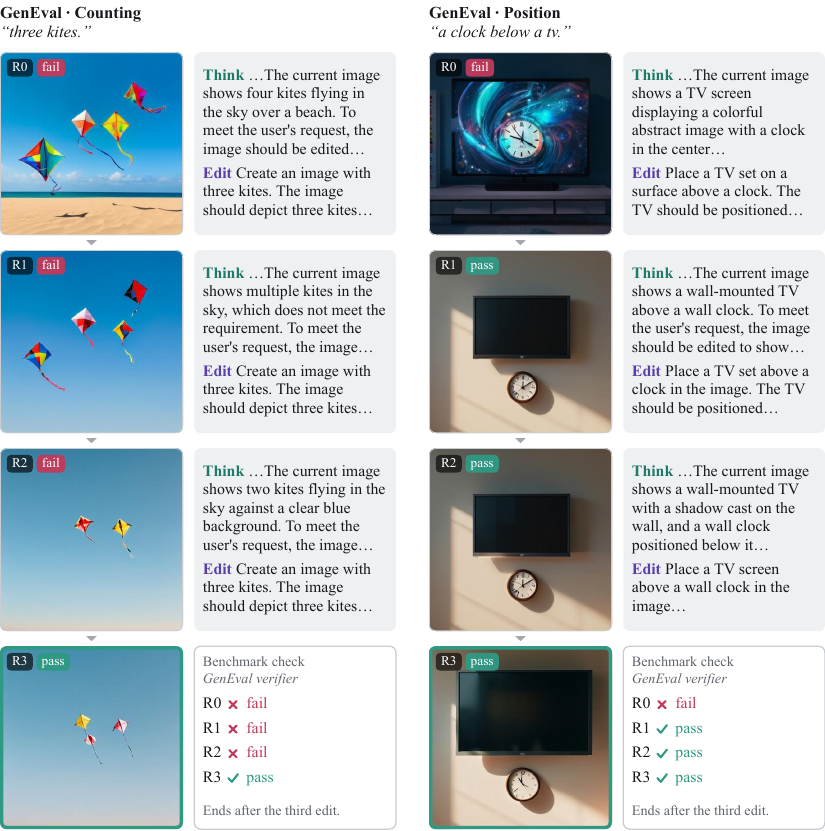}
\caption{\textbf{Successful repairs.} Each column reads top to bottom: the image, the model's verbatim reflection, and the image it renders next. Frames mark the benchmark verdict. Cases are selected.}
\label{fig:qual_success_1}
\end{figure}

\begin{figure}[p]
\centering
\includegraphics[width=\linewidth]{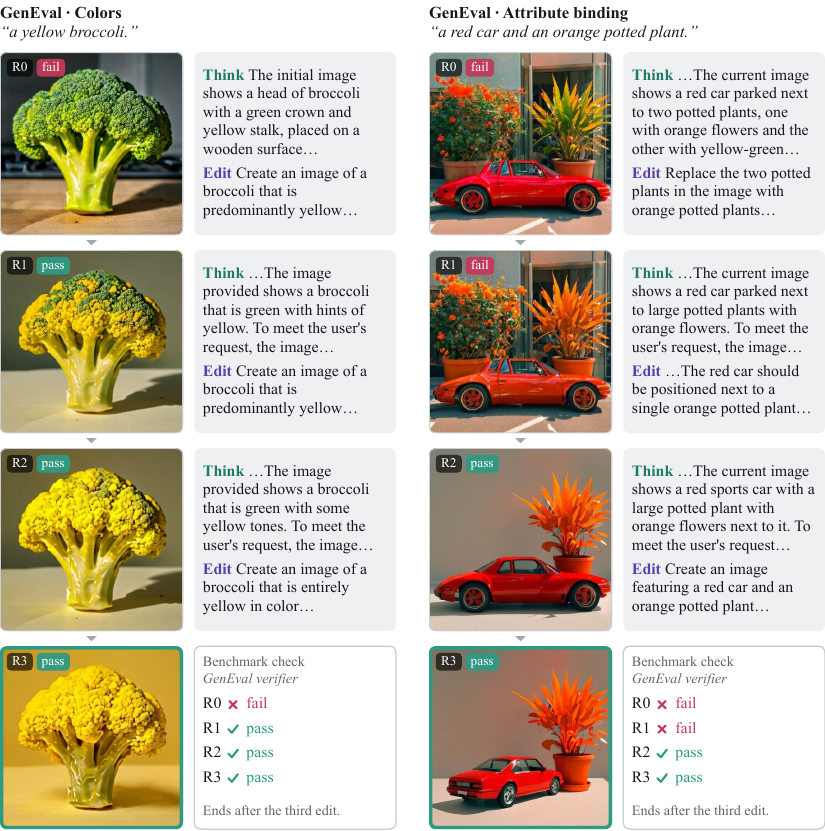}
\caption{\textbf{Successful repairs} (continued); layout as in Figure~\ref{fig:qual_success_1}.}
\label{fig:qual_success_2}
\end{figure}

\begin{figure}[p]
\centering
\includegraphics[width=\linewidth]{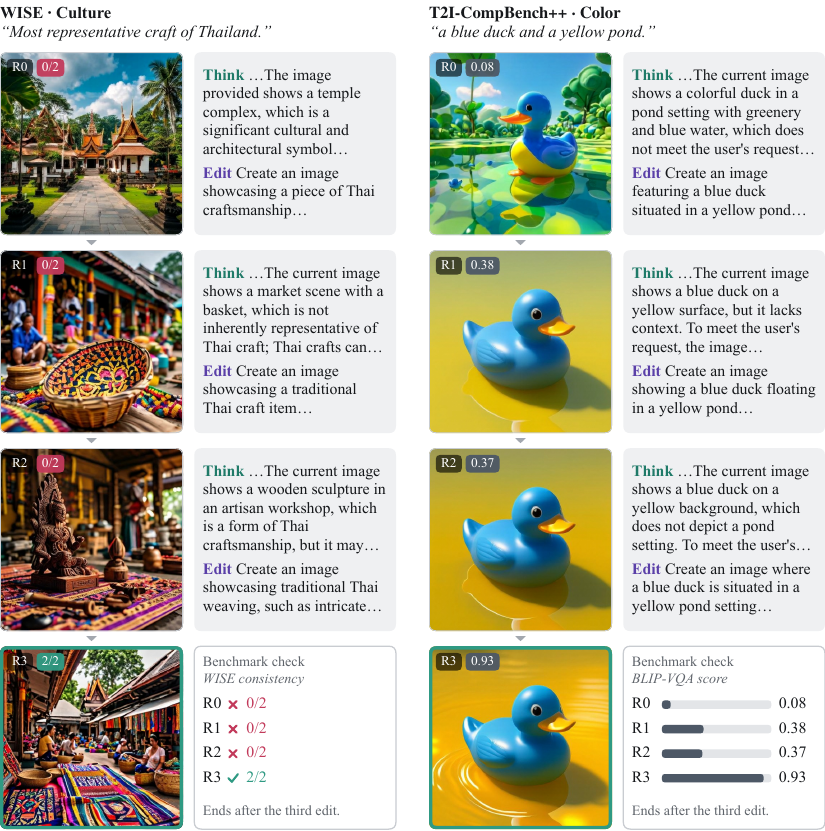}
\caption{\textbf{Successful repairs} (continued); layout as in Figure~\ref{fig:qual_success_1}.}
\label{fig:qual_success_3}
\end{figure}

\begin{figure}[p]
\centering
\includegraphics[width=\linewidth]{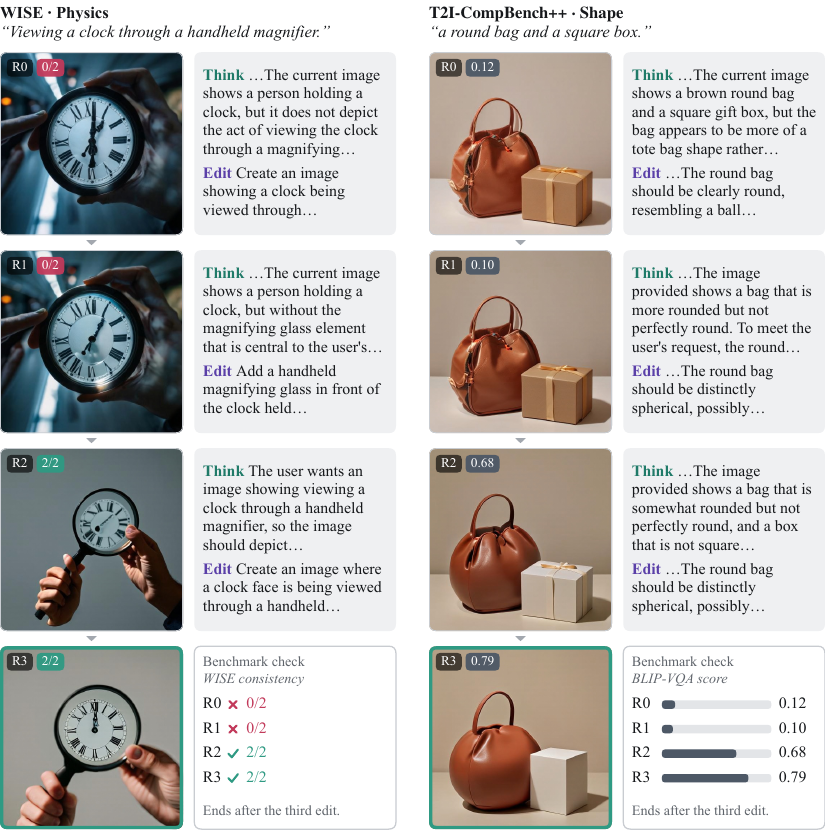}
\caption{\textbf{Successful repairs} (continued); layout as in Figure~\ref{fig:qual_success_1}.}
\label{fig:qual_success_4}
\end{figure}

\begin{figure}[p]
\centering
\includegraphics[width=\linewidth]{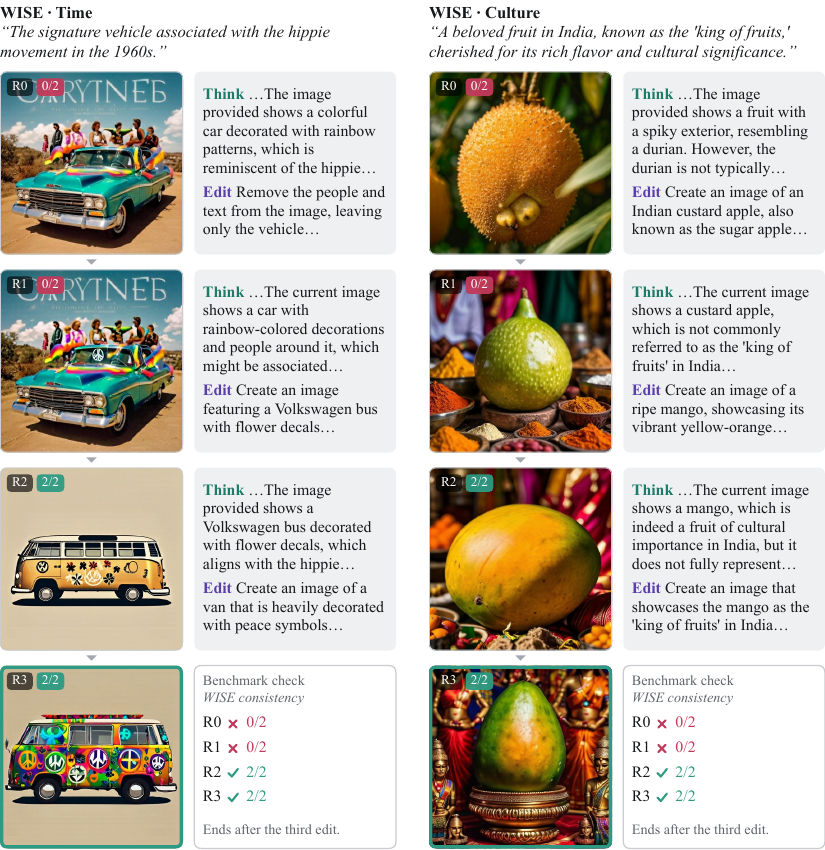}
\caption{\textbf{Successful repairs} (continued); layout as in Figure~\ref{fig:qual_success_1}.}
\label{fig:qual_success_5}
\end{figure}

\begin{figure}[p]
\centering
\includegraphics[width=\linewidth]{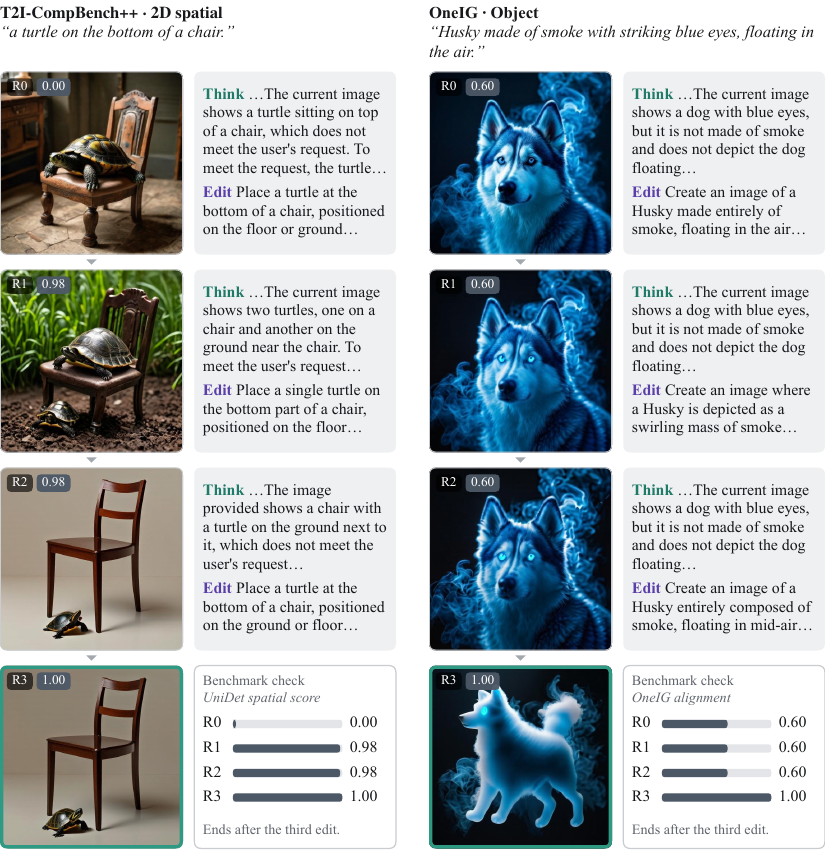}
\caption{\textbf{Successful repairs} (continued); layout as in Figure~\ref{fig:qual_success_1}.}
\label{fig:qual_success_6}
\end{figure}

\begin{figure}[p]
\centering
\includegraphics[width=\linewidth]{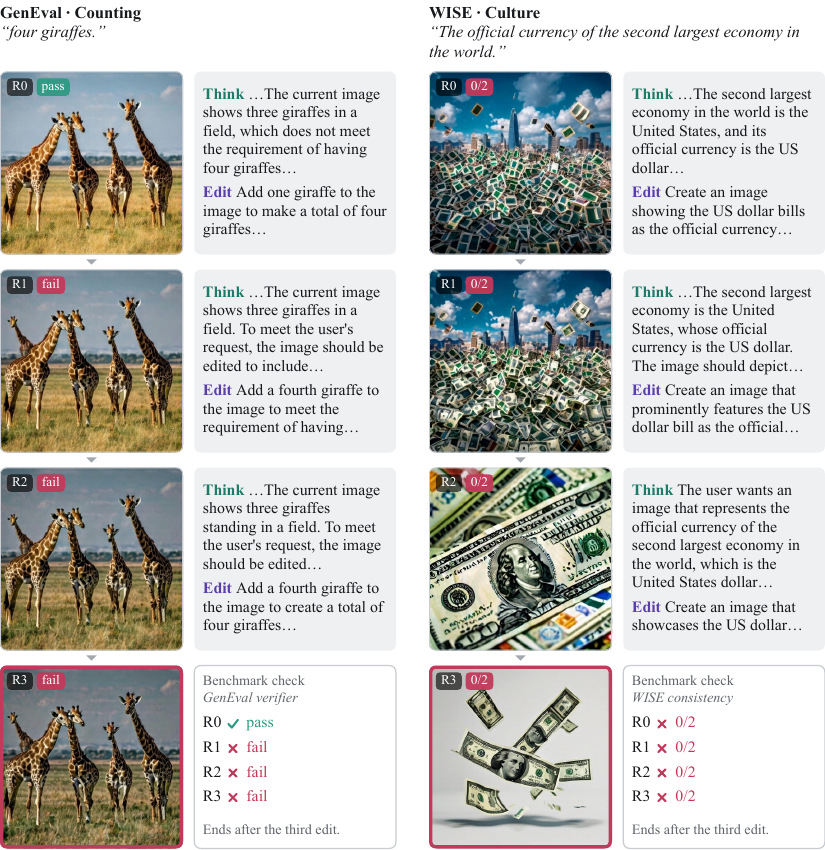}
\caption{\textbf{Failure cases.} Failures caused by the reflection. Left: the first image already shows four giraffes and passes, but the reflection counts them as three and asks for one more; it repeats ``three giraffes'' in every later round while the image holds five. Right: the reflection states that the second largest economy is the United States, so all three edits render US dollars instead of the Chinese yuan. Frames mark the benchmark verdict.}
\label{fig:qual_failure}
\end{figure}

Figures~\ref{fig:qual_success_1}--\ref{fig:qual_success_6} show further successful repairs on all four benchmarks, and Figure~\ref{fig:qual_failure} shows two failures caused by the reflection itself.

\FloatBarrier

\FloatBarrier
\section{Stopping Behavior under Terminal Rewards}
\label{app:stopping}

The reflection protocol lets the policy end a trajectory with DONE at any
round. How the terminal round is priced determines when the policy stops;
we compare two prices on the same parent, prompt stream, and seed, both at
500 updates.

\paragraph{Penalty on a wrong DONE (primary reward).}
A DONE emitted on a verifier-incorrect image costs $0.5$; a DONE on a
correct image earns nothing beyond the terminal quality. The gains in Tables~\ref{tab:geneval}, \ref{tab:main}, and~\ref{tab:ablation} are
realized in this regime, with accuracy rising through rounds one, two, and
three.

\paragraph{Adding a correct-DONE bonus.}
Adding a bonus for a DONE on a correct image, together with a $-0.5$ action
penalty on editing an already-correct image, lowers the confidence at which
stopping breaks even from 1.0 to 0.5. Stopping becomes cheap, and the policy
learns to stop prematurely: it declares DONE on 112 incorrect images, which
make up 112 of its 116 final failures, and final accuracy falls from 82 to
79. At a break-even of 0.5, a stop pays off whenever the image is as likely
wrong as right, so the policy gives up on images it could still repair.

\paragraph{Summary.}
A correct-DONE bonus makes stopping cheap and turns reflection into premature
acceptance of failed images. The wrong-DONE penalty keeps the policy repairing
and gives the highest final accuracy; \method{} therefore uses the penalty
alone.

\section{Per-Family Learning on GenEval}
\label{app:families}

\begin{figure}[ht]
\centering
\includegraphics[width=\linewidth]{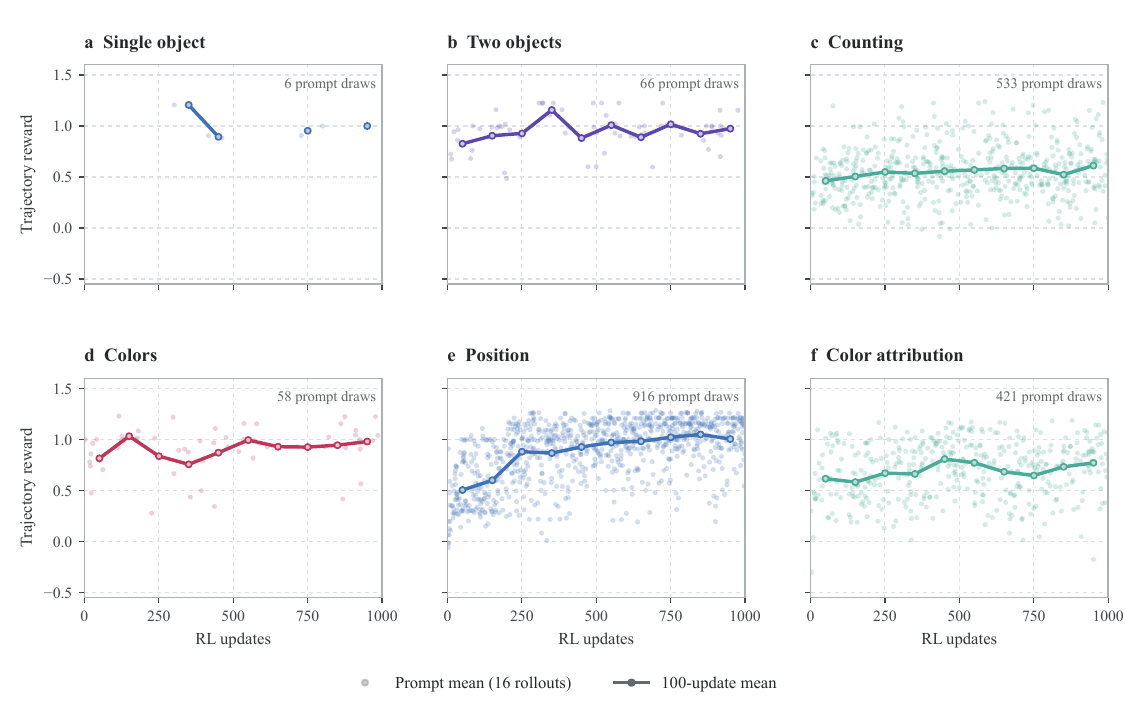}
\caption{\textbf{Training reward by GenEval family.} Light points are the
mean whole-trajectory reward of the 16 rollouts for one prompt; lines are
100-update means. Prompt draws per family follow the training pool
composition; single object has six draws and its line is left open.}
\label{fig:family_reward}
\end{figure}

The six GenEval families learn at different rates, and the training reward
predicts the held-out gain (Figure~\ref{fig:family_reward}). Position
rises most, from about 0.5 to 1.0 over the run, and its GenEval accuracy
rises from 55 to 89. Color attribution and colors rise moderately and gain
17 and 6 points. Two objects starts high and gains 12 points; single object
is at ceiling. Counting is the one family whose training reward stays flat
across 1{,}000 updates, and its GenEval accuracy is unchanged at 67.5: the
verifier requires an exact count, and the edits the policy learns, which
add, remove, recolor, and reposition objects, do not yet move the count
reliably. Counting therefore identifies the next training target for
reflection, count-aware edits, where the same reward and protocol apply
without change.

\section{Inference System Prompt}
\label{app:systemprompt}

The system prompt used by the native inference driver is reproduced below.

\begin{quote}
\small
You are an image generation, editing, and verification agent operating over an ordered visual trajectory.

The original user request remains the final objective throughout the trajectory. A T2I trajectory begins with no source image; an edit trajectory begins with a given source image. Later visible images are the current intermediate state.

At every reasoning turn, output exactly one assistant response: use one \textless{}think\textgreater{} block with these exact tags: [CURRENT\_ROUND], [SCORE], [ACTION], [THINKING], and [SOURCE\_IMAGE], then place exactly one [EDIT] field immediately after \textless{}/think\textgreater{} in the same response.

Use [SOURCE\_IMAGE] None before initial T2I generation, given for an external edit source, or Image \#N for a generated intermediate image. Use [ACTION] edit with one concrete [EDIT] instruction to generate or revise the next image. Use [ACTION] done only when the complete original request is visibly satisfied, with [EDIT] None.

For a planned progression, execute only the currently due milestone. Constraints assigned to future milestones are intentionally pending, not model failures or defects in the current image. Preserve completed milestones while advancing the next one. Call a constraint failed only when it was due and is visibly incorrect.

Judge only visible evidence, preserve unrelated content for edits, and do not invent corruption, rollback state, or hidden success.

For a planned progression, begin first-round [THINKING] with the complete compact milestone plan, then execute only the currently due milestone. Keep that plan in the persistent history for later verification turns.

The [SCORE] field is always written on the fixed protocol scale as N/10, where N is an integer from 0 to 10 and 10 means the original request is fully satisfied by the current image. Never use another denominator, a percentage, or a bare number.

If the current image already satisfies the complete original request, emit [ACTION] done with [EDIT] None. Never emit [ACTION] edit with an empty, None, or placeholder [EDIT] payload; that is a done decision.

\end{quote}

\end{document}